\documentclass{article}
\usepackage[margin=1in]{geometry}
\usepackage[T1]{fontenc}
\usepackage[utf8]{inputenc}
\usepackage{authblk} 
\usepackage{textcomp}
\usepackage{lmodern}
\usepackage{amsmath, amssymb}
\usepackage{graphicx}
\usepackage[numbers]{natbib}
\usepackage{hyperref}
\usepackage{tabularx, booktabs, array}
\usepackage{longtable}
\usepackage{multirow}
\usepackage{makecell}
\newcolumntype{L}{>{\raggedright\arraybackslash}X}
\usepackage{placeins}
\usepackage{float}
\usepackage{titlesec}
\titleformat{\section}{\normalfont\fontsize{12}{14}\bfseries\sffamily}{\thesection}{1em}{}
\titleformat{\subsection}{\normalfont\fontsize{12}{14}\bfseries\sffamily}{\thesubsection}{1em}{}
\usepackage{caption}
\title{\bfseries A Synergistic Framework of Nonlinear Acoustic Computing and Reinforcement Learning for Real-World Human-Robot Interaction}

\author[1]{Xiaoliang Chen\thanks{Corresponding author: chenxiaoliang@soundai.com}}
\author[1]{Xin Yu}
\author[1]{Le Chang}
\author[1]{Yunhe Huang}
\author[1]{Jiashuai He}
\author[1]{Shibo Zhang}
\author[1]{Jin Li}
\author[1]{Likai Lin}
\author[1]{Ziyu Zeng}
\author[1]{Xianling Tu}
\author[1]{Shuyu Zhang}

\affil[1]{SoundAI Technology Co., Ltd.}

\date{}

\begin{document}


\maketitle

\begin{abstract}
This paper introduces a novel framework integrating nonlinear acoustic computing and reinforcement learning to enhance advanced human-robot interaction under complex noise and reverberation. Leveraging physically informed wave equations (e.g., Westervelt, KZK), the approach captures higher-order phenomena such as harmonic generation and shock formation. By embedding these models in a reinforcement learning-driven control loop, the system adaptively optimizes key parameters (e.g., absorption, beamforming) to mitigate multipath interference and non-stationary noise. Experimental evaluations—covering far-field localization, weak signal detection, and multilingual speech recognition—demonstrate that this hybrid strategy surpasses traditional linear methods and purely data-driven baselines, achieving superior noise suppression, minimal latency, and robust accuracy in demanding real-world scenarios. The proposed system demonstrates broad application prospects in AI hardware, robot, machine audition, artificial audition, and brain-machine interfaces.
\end{abstract}


\section{Introduction}

With the rapid growth of artificial intelligence (AI), especially in the fields of deep learning and reinforcement learning, acoustic computing and human-robot interaction have entered a phase of unprecedented expansion and innovation. Recent advances in audio signal processing, speech recognition, and targeted signal enhancement have not only boosted computational efficiency but also redefined the scope and depth of human-machine communication in diverse application domains. Despite notable successes stemming from conventional acoustic methods, these linear or quasi-linear approaches often exhibit critical shortcomings in real-world settings. Specifically, far-field speech processing, detection of weak acoustic signals, and sophisticated noise suppression remain open problems—particularly in dynamic, cluttered, or reverberant environments \cite{rabiner1993fundamentals}.

Traditional acoustic solutions predominantly rely on physical models—such as geometric room models and linear wave propagation equations—to describe sound propagation phenomena like source localization, reflection cancellation, and multipath mitigation. Although these techniques have proven effective in controlled or moderately noisy settings, the acoustic fields in real scenarios frequently violate the assumptions of linearity and stationarity. As acoustic complexity grows—due to strong reverberation, moving noise sources, or nonstationary interference—classical models fail to adapt quickly enough or handle higher-order effects accurately. In highly reverberant or enclosed spaces with strong reflections, classical signal processing pipelines require extensive tuning, often leading to suboptimal real-time performance. Meanwhile, strong noise sources, such as industrial machinery or multi-speaker chatter, can severely degrade speech intelligibility. Motivated by these limitations, more flexible approaches that embed intelligence within the processing pipeline are increasingly sought. Reinforcement learning applied to acoustic problems thus holds considerable promise: it enables model-based or model-free adaptive control that can dynamically optimize filter coefficients, beamforming weights, and other parameters in response to changing conditions.

Against this backdrop, this paper proposes an innovative framework that integrates \emph{nonlinear acoustic computing} with \emph{reinforcement learning}, creating a tightly coupled system capable of operating reliably in harsh acoustic environments. Nonlinear acoustic theories, derived from fundamental wave equations such as the Westervelt and Khokhlov-Zabolotskaya-Kuznetsov (KZK) equations, capture higher-order phenomena (harmonic generation, shock formation, waveform distortion, etc.) beyond the reach of simple linear models. These effects become pronounced when acoustic pressures are high or environmental boundary conditions induce notable nonlinearities, including strong reflections or concentrated energy focusing. Reinforcement learning augments these physically informed models by facilitating real-time parameter tuning—such as automatically adjusting attenuation coefficients and source localization strategies—to mitigate factors like multipath interference and unpredictable noise. Together, these components form an adaptive control loop, continuously refining both the physics-based model parameters and the signal processing front-end for robust performance.

One of the central challenges in developing robust acoustic solutions lies in balancing the interpretability of physics-based modeling with the adaptivity of data-driven methods. Although modern deep neural networks have demonstrated exceptional performance in speech enhancement and recognition tasks, they often demand vast labeled datasets and exhibit limited extrapolation to unusual or unseen acoustic conditions. Conversely, physics-based models deliver interpretable insights into wave propagation but typically lack the mechanisms to adapt in real time to rapidly changing environments. Melding these perspectives is, therefore, both an intellectual and practical necessity. Our hybrid framework systematically fuses explicit domain knowledge—encoded through the Westervelt/KZK equations—with the learning-driven flexibility of reinforcement learning. This synergy leverages the best of both worlds, enhancing generalization without losing the underlying physical consistency that fosters interpretable and reliable performance.

Nonlinear acoustic models—exemplified by the Westervelt and KZK formulations—hold particular promise for high-intensity or large-scale applications, where wave nonlinearities manifest in ways that linear approximations cannot capture. By embedding these nonlinear equations into an end-to-end pipeline, we exploit physically grounded terms for diffraction, harmonic generation, and absorption, thus enabling advanced noise rejection and directional filtering. Within this pipeline, reinforcement learning serves as a meta-optimizer: it incrementally calibrates beamforming weights, filter coefficients, and boundary reflection parameters based on continuous feedback. Such a scheme alleviates the need for constant human supervision or meticulous hand-tuning across varying acoustic environments—industrial factories, vehicular cabins, large auditoriums, and beyond. The payoff is enhanced source localization accuracy, low-latency speech recognition, and more robust detection of weak or occluded signals.

The primary contributions of this work can be summarized as follows:
\begin{enumerate}
\item \textbf{Integration of Nonlinear Models and Reinforcement Learning:} We introduce a novel scheme that leverages the predictive power of nonlinear wave equations within an adaptive control context. By combining physically grounded modeling with deep reinforcement learning, our framework dynamically adapts to nonstationary or reverberant environments in real time.
\item \textbf{Joint Optimization of Physics-Informed and Data-Driven Methods:} We devise a co-optimization methodology that balances the interpretability of wave propagation models against the noise robustness and flexibility of advanced neural architectures. This synergistic approach significantly boosts speech feature extraction, denoising, and far-field source localization capabilities.
\item \textbf{Validation Through Extensive Experiments:} We present a thorough experimental campaign in challenging acoustic setups, including environments with persistent multipath reflections, high-intensity noise sources, and fast-changing interference. Compared to state-of-the-art baselines, our framework consistently achieves improved speech recognition accuracy, faster computational throughput, and lower latency, opening new avenues for real-world deployments.
\end{enumerate}

\begin{figure}[htbp]
  \centering
  \includegraphics[width=\linewidth]{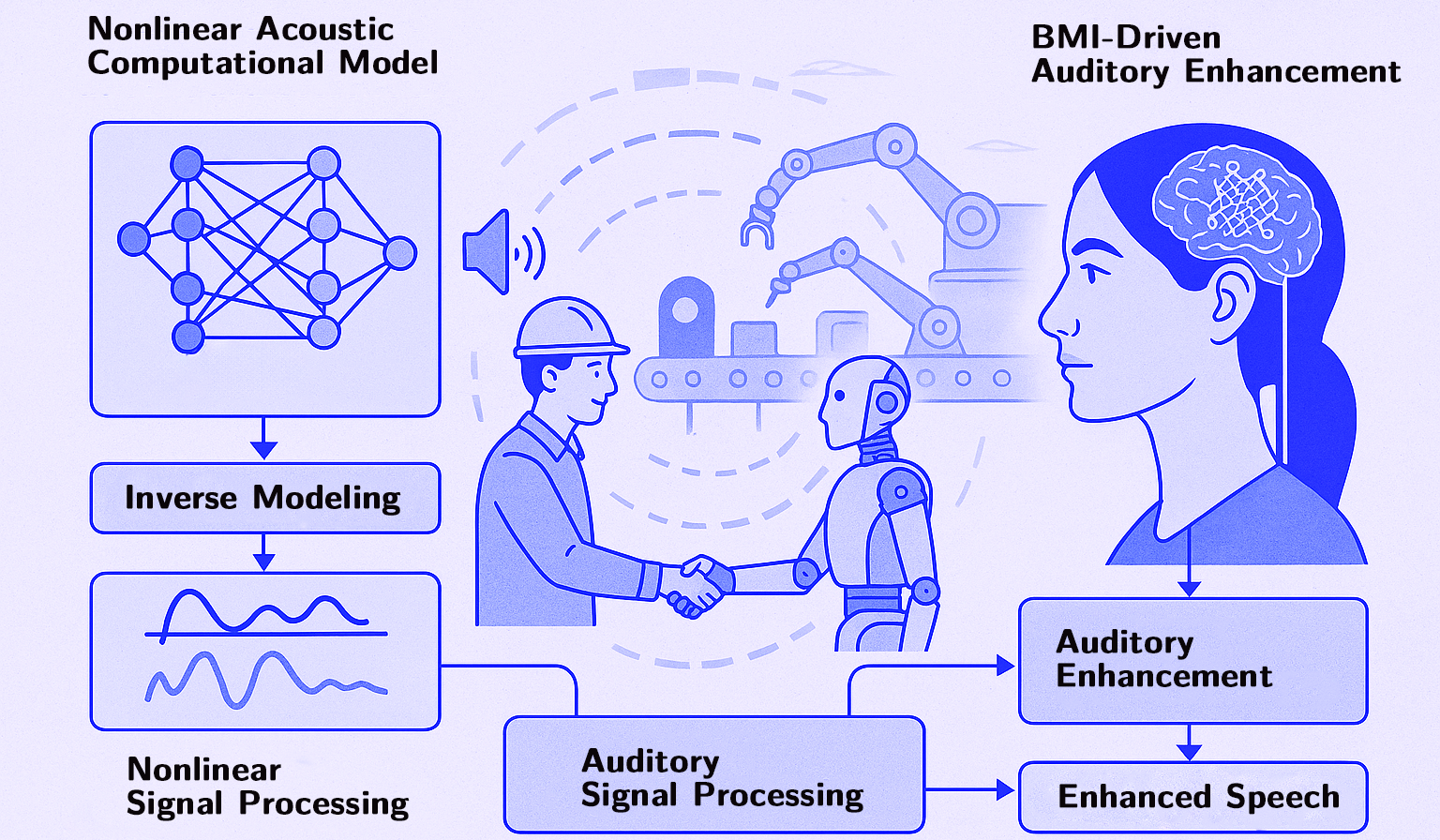}
  \vspace{-15pt}
  \caption{Nonlinear Acoustic Computational Model and Auditory Enhancement Framework Driven by Brain-Machine Interface}
  \label{fig:nonlinear_model_and_auditory_framework}
\end{figure}

Figure~\ref{fig:nonlinear_model_and_auditory_framework} offers an overview of the proposed system. As shown:

\textbf{Left:} A physics-informed nonlinear acoustic computational model underpins the entire pipeline, employing reinforcement learning to estimate and update the key parameters from the Westervelt/KZK equations. These adaptive coefficients feed downstream signal processing blocks, bridging the gap between pure theoretical approximations and real-world acoustic data.

\textbf{Center:} A real-time auditory signal processing module encompasses adaptive filtering, source separation, and echo cancellation. Operating in tandem with the updated nonlinear model parameters, this subsystem addresses multiple acoustic challenges—ranging from abrupt changes in environmental noise to multi-speaker overlaps—empowering advanced human-robot interaction in noisy or highly reverberant spaces.

\textbf{Right:} In parallel, we incorporate a BMI-driven auditory enhancement unit. Neural and cortical signals are decoded and mapped onto parameters for dynamic speech reconstruction, offering individualized adjustments in intelligibility and perceptual quality. This integration lays the groundwork for next-generation brain-machine interfaces that combine advanced signal processing with user-specific customization.

\FloatBarrier

The rest of this paper is organized as follows. In Section~2, we describe the theoretical underpinnings of nonlinear acoustic computing, focusing on the derivations and practical aspects of the Westervelt and KZK equations, and propose a new reinforcement-learning-driven architecture for real-time parameter updating. Section~3 details the core algorithms, including AI-based noise reduction and echo cancellation (Section~3.1), a full-scenario sensor-array source localization approach (Section~3.2), a multilingual speech interaction pipeline capable of handling 66 languages (Section~3.3), and a multimodal large-model system supporting integrated acoustic-vision tasks (Section~3.4). Section~4 describes our experimental setup and results, highlighting metrics such as speech recognition accuracy, PESQ/MOS-LQO for noise suppression, real-time factor (RTF), and emotion/event detection, and comparing these against relevant baselines. Section~5 surveys various application scenarios, including AI headphones and smart speakers (Section~5.1), adaptive AI hearing aids (Section~5.2), multifunctional microphones and robot audition (Section~5.3), brain-machine interface hearing evaluations (Section~5.4), and novel in-vehicle acoustic systems (Section~5.5). Finally, Section~6 concludes the paper, outlining promising future directions such as multimodal fusion, real-world experience, and the continuing evolution of next-generation human-robot interaction platforms.

\section{Theoretical Foundations and Innovative Framework}
\label{sec:theoretical_framework}

The robust performance of far-field acoustic interaction systems—covering applications such as robotic dialogue, speech recognition, voice cloning, and complex environment modeling—critically hinges on accurate physical modeling of wave propagation and adaptive algorithmic strategies. Traditional linear acoustic models often fail to capture essential phenomena such as harmonic generation, waveform distortion, and phase shifts, which arise in high-intensity or reverberant media. To address these limitations, this work integrates rigorous nonlinear acoustic formulations with state-of-the-art machine learning techniques.

\subsection{Nonlinear Acoustic Modeling}

Nonlinear acoustic models extend classical linear theory by incorporating medium compressibility and boundary interactions that induce amplitude-dependent effects. More specifically, when acoustic pressure amplitudes become sufficiently large or when the propagation medium/media has strong boundary reflections, linear approximations can no longer accurately capture the underlying physics\cite{zuwen1999nonlinear}. 

Starting from the compressible Navier-Stokes equations under the assumptions of irrotational and lossless flow, one can obtain various forms of nonlinear acoustic wave equations through perturbation expansions. Let the total acoustic pressure be 
\[
  p = p_0 + \epsilon p_1 + \epsilon^2 p_2 + \cdots,
\]
where \(p_0\) is the ambient (static) pressure, and \(\epsilon\) is a small parameter characterizing the relative amplitude of the acoustic perturbation. Retaining terms up to \(\mathcal{O}(\epsilon^2)\) (i.e., second-order terms in \(\epsilon\)) and neglecting viscous losses yields the so-called Westervelt Equation, discussed in detail below.

\subsubsection{The Westervelt Equation}
\label{subsubsec:westervelt}

By applying the second-order perturbation expansion to the governing Navier-Stokes equations under the assumptions stated above, one first arrives at an intermediate form:
\begin{align}
  \frac{\partial^2 p_1}{\partial t^2} 
    \;-\; c^2 \nabla^2 p_1 
    &= -\,\epsilon \,\frac{\beta}{\rho_0\,c^2} \,\frac{\partial^2 (p_1^2)}{\partial t^2},
  \label{eq:westervelt_perturbation}
\end{align}
where 
\begin{itemize}
  \item \(\rho_0\) is the ambient density,
  \item \(c\) is the small-signal (linear) sound speed,
  \item \(\beta = 1 + \tfrac{B}{2A}\) is the coefficient of nonlinearity, with \(\tfrac{B}{A}\) often arising from the equation of state for the medium.
\end{itemize}

From Equation~\eqref{eq:westervelt_perturbation}, by normalizing and redefining parameters (absorbing \(\epsilon\) into the new pressure variable) and dropping higher-order dissipative terms, one obtains the canonical Westervelt Equation:
\begin{equation}
  \frac{\partial^2 p}{\partial t^2} 
    \;-\; c^2 \nabla^2 p 
    = \alpha \frac{\partial^2 (p^2)}{\partial t^2},
  \label{eq:westervelt}
\end{equation}
where \(\alpha \propto \frac{\beta}{\rho_0\,c^2}\) encapsulates the medium’s nonlinear response. 

~

Equation~\eqref{eq:westervelt} (the Westervelt Equation) captures several important phenomena in high-amplitude or strongly nonlinear acoustic fields:
\begin{itemize}
  \item \emph{Second-harmonic generation:} The quadratic term on the right-hand side acts as a source for higher harmonics, notably doubling the fundamental frequency.
  \item \emph{Waveform steepening and shock formation:} As amplitude grows, the wavefront can steepen, ultimately leading to shock waves if the propagation distance is large enough.
  \item \emph{Amplitude-dependent attenuation:} Although not explicitly modeling linear or thermoviscous absorption, the nonlinearity effectively introduces additional amplitude-dependent propagation effects.
\end{itemize}
In many practical contexts—such as biomedical ultrasound, high-intensity focused ultrasound, and certain robotic audition scenarios—these nonlinear behaviors cannot be ignored when attempting to accurately model wave propagation.

\subsubsection{Derivation of the KZK Equation}
\label{subsubsec:kzk}

For quasi-planar, directional beams in the far field, one must account for nonlinearity, diffraction, and absorption. The Khokhlov-Zabolotskaya-Kuznetsov (KZK) equation is derived by applying the paraxial approximation (\(\partial/\partial z \ll \partial/\partial r\)) and using a retarded time frame \(\tau = t - z/c\). This leads to the KZK equation:
\begin{equation}
  \frac{\partial^2 p}{\partial z\,\partial \tau}
    =  \frac{c}{2}\,\nabla_\perp^2 p \;+\;  \frac{\beta_{\text{NL}}}{2\,\rho_0\,c^3}\,\frac{\partial^3 (p^2)}{\partial \tau^3}
      \;+\; \frac{\delta}{2\,c^3}\,\frac{\partial^3 p}{\partial \tau^3},
  \label{eq:kzk_full}
\end{equation}
where \(\nabla_\perp^2\) represents the transverse Laplacian, \(\beta_{\text{NL}}\) is the coefficient of nonlinearity (e.g., \(1 + B/2A\)), \(\rho_0\) is the ambient density, \(c\) is the sound speed, and \(\delta\) models thermoviscous absorption.

By neglecting absorption (\(\delta = 0\)), a simplified, lossless form is obtained:
\begin{equation}
  \frac{\partial^2 p}{\partial z\,\partial \tau}
    = \frac{c}{2}\,\nabla_\perp^2 p
      \;+\; \frac{\beta_{\text{NL}}}{2\,\rho_0\,c^3}\,\frac{\partial^3 (p^2)}{\partial \tau^3}.
  \label{eq:kzk_simple}
\end{equation}

For an axisymmetric beam in cylindrical coordinates \((r, z, \tau)\), where the transverse Laplacian is \(\nabla_\perp^2 p = \frac{1}{r}\frac{\partial}{\partial r}\left(r\frac{\partial p}{\partial r}\right)\), the lossless KZK equation (from Eq.~\eqref{eq:kzk_simple}) can be written explicitly as:
\begin{equation}
  \frac{\partial^2 p}{\partial z\,\partial \tau}
    = \frac{c}{2r}\,\frac{\partial}{\partial r}\!\left(r\,\frac{\partial p}{\partial r}\right)
      \;+\; \frac{\beta_{\text{NL}}}{2\,\rho_0\,c^3}\,\frac{\partial^3 (p^2)}{\partial \tau^3}.
  \label{eq:kzk} 
\end{equation}

This form explicitly shows how nonlinearity (via \(\frac{\partial^3 (p^2)}{\partial \tau^3}\)) and diffraction (via the radial derivative term) combine to shape beam patterns and harmonic content in axisymmetric systems.
The KZK Equation (in its general forms \eqref{eq:kzk_full}, \eqref{eq:kzk_simple}, or specialized form \eqref{eq:kzk}) can be adapted for different boundary conditions and is extensively used in modeling focused acoustic holography, and related far-field scenarios \cite{zuwen1999nonlinear}.

\subsection{Integration with Machine Learning}

While the aforementioned nonlinear models (Equations~\eqref{eq:westervelt} and \eqref{eq:kzk}) provide a rigorous physical foundation, real-world deployment demands efficient parameter estimation and dynamic adaptation to diverse acoustic environments. We propose a dual-pathway framework—integrating deep learning and reinforcement learning—to achieve both high fidelity and robust adaptability.

\subsubsection{Deep Learning Synergy}

We employ deep neural networks (e.g., CNNs, RNNs, LSTMs \cite{hori2017multi,lecun2015deep,graves2013speech}) to extract high-dimensional acoustic features and perform real-time denoising and parameter tracking:
\begin{enumerate}
  \item \textbf{Feature Extraction:} Multi-scale convolutional filters capture time-frequency patterns that characterize nonlinear distortion and reverberant tails.
  \item \textbf{Parameter Estimation:} A learned regression head maps hidden activations to adaptive estimates of \(\alpha\), \(\delta\), and boundary impedances in Equations~\eqref{eq:westervelt}--\eqref{eq:kzk}.
  \item \textbf{Denoising \& Separation:} Mask-based spectral filtering isolates target speech, yielding clean signals for the physics-based propagators.
  \item \textbf{Hardware Acceleration:} Model quantization and tensor optimizations enable sub-millisecond inference on edge AI processors.
\end{enumerate}
During deployment, the network’s output continuously refines the nonlinear model parameters, forming a feedback loop to reduce mismatches between nominal physics assumptions and real-world acoustics.

\subsubsection{Reinforcement Learning Fusion}

To further enhance adaptability in time-varying, multipath-rich environments, we embed a reinforcement learning (RL) agent\cite{latif2023survey} to optimize end-to-end performance (e.g., word error rate, signal-to-noise ratio):
\begin{itemize}
  \item \textbf{State} \(\mathbf{s}_t\): Encodes recent acoustic observations, model parameter estimates, and recognition confidence scores.
  \item \textbf{Action} \(\mathbf{a}_t\): Chooses incremental adjustments to propagation coefficients, filter gains, and beamforming weights.
  \item \textbf{Reward} \(r_t\): A composite metric balancing recognition accuracy, computational latency, and energy consumption.
\end{itemize}
We employ proximal policy optimization (PPO) to learn the policy \(\pi(\mathbf{a}_t|\mathbf{s}_t)\):
\begin{equation}
  L^{\mathrm{PPO}}(\theta) 
   = \mathbb{E}_t \Bigl[
      \min\!\Bigl(r_t(\theta)\hat{A}_t,\;
      \mathrm{clip}\bigl(r_t(\theta),1-\epsilon,1+\epsilon\bigr)\hat{A}_t\Bigr)
     \Bigr],
  \label{eq:ppo}
\end{equation}
where \(r_t(\theta)\) is the likelihood ratio and \(\hat{A}_t\) the advantage estimate. Through continuous interaction, the RL agent learns to trade off:
\begin{itemize}
  \item \emph{Noise suppression} vs.\ \emph{latency} for real-time robotics,
  \item \emph{Echo cancellation} vs.\ \emph{beamwidth} in multi-speaker settings,
  \item \emph{Energy efficiency} vs.\ \emph{accuracy} in battery-powered devices.
\end{itemize}

\begin{figure}[htbp]
  \centering
  \includegraphics[width=\linewidth]{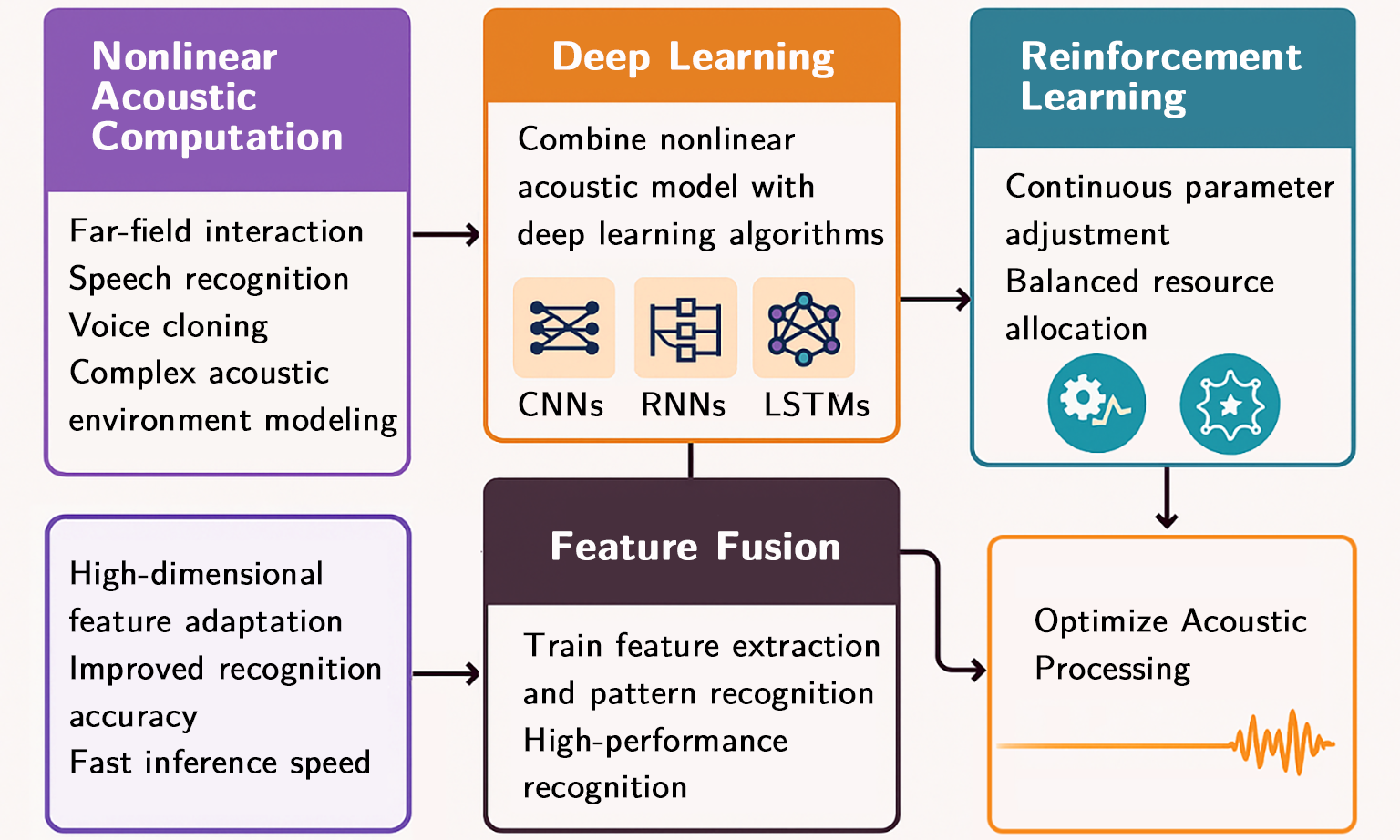}
  \caption{Fusion Framework of Nonlinear Acoustic Computation and Machine Learning}
  \label{fig:fusion_framework}
\end{figure}

Figure~\ref{fig:fusion_framework} illustrates the tight coupling between physics-based wave propagation and learning-driven modules, enabling end-to-end optimization of far-field voice interaction systems. By combining the interpretability of nonlinear acoustic equations (Equations~\eqref{eq:westervelt} and \eqref{eq:kzk}) with the adaptability of deep and reinforcement learning, the proposed framework aims to deliver robust performance across a broad range of real-world scenarios.

\section{Algorithms, Models, and Technical Indicators}
\label{sec:algorithms_models_indicators}

Over the last few decades, acoustic computing technologies have made tremendous strides, particularly in the domain of far-field acoustic interaction. Multiple cutting-edge algorithms \cite{ma2004foundation} have surfaced to tackle the considerable difficulties associated with sound wave propagation in highly intricate environments. Far-field acoustic interaction encompasses an ecosystem of acoustic models and signal processing techniques intended to enhance speech recognition accuracy under notoriously harsh conditions such as strong noise, severe reverberation, and pronounced signal attenuation. This area has often been deemed the “holy grail” of intelligent audio solutions, reflecting both the complexity of the domain and the potential gains from successful implementations.

\subsection{AI Noise Reduction and Echo Cancellation Algorithms}
Noise reduction and echo cancellation are commonly viewed as the “bread and butter” of far-field acoustic interaction. Without robust solutions to mitigate the detrimental effects of ambient and reverberant noise, speech recognition systems remain highly vulnerable to performance drops \cite{hershey2016deep}. One compelling approach to echo cancellation in extremely noisy environments (e.g., signal-to-noise ratios as low as -5 dB) is a subband decomposition-based lightweight algorithm. By splitting the broadband acoustic signal into multiple frequency bands and handling each band’s echo components independently \cite{loizou2007speech}, this technique not only decreases computational load but also preserves signal fidelity and echo suppression capability \cite{chen2010high}.

To shed more light on the mechanics of subband decomposition, let the original time-domain signal be \( x(n) \). We decompose \( x(n) \) into \( M \) subbands \( x_m(k) \) via an analysis filter bank \( H_m(e^{j\omega}) \), yielding
\begin{equation}
  x_m(k) = \sum_{n=0}^{\infty} x(n)\, h_m(L\,k-n),
  \label{eq:subband_decomp}
\end{equation}
where \( L \) is the decimation factor. Each subband \( x_m(k) \) can be processed with adaptive echo cancellation filters \( W_m \) to remove echoes band by band:
\begin{equation}
  e_m(k) = x_m(k) - W_m(k) * d_m(k),
  \label{eq:subband_ec}
\end{equation}
where \( d_m(k) \) denotes the desired (echo) component in the \( m \)-th band. Summation of the processed subbands through a synthesis filter bank reconstructs the time-domain signal with minimized echoes. Experimental results show that this algorithm significantly boosts the wake-up rate to 96\% while simultaneously cutting power consumption by 40\%, providing a “best-of-both-worlds” solution for on-device intelligence.

In parallel, researchers have also pulled out all the stops to tackle multimodal noise separation. A particularly notable strategy integrates voiceprints, lip movements, and thermal maps within a single system. This multifaceted design can elevate the signal-to-noise ratio (SNR) by as much as 12 dB in industrial environments experiencing noise levels around 100 dB \cite{chen2023challenges}. Given its resilience and stability, the approach has found success at large-scale events like the Winter Olympics and in spacious venues such as bustling convention centers. Not only does it furnish stable and clear speech recognition, but it also supplies tangible evidence that fusing multiple modalities can be a powerful “silver bullet” against severe environmental noise.

\begin{figure}[htbp]
  \centering
  \includegraphics[width=\linewidth]{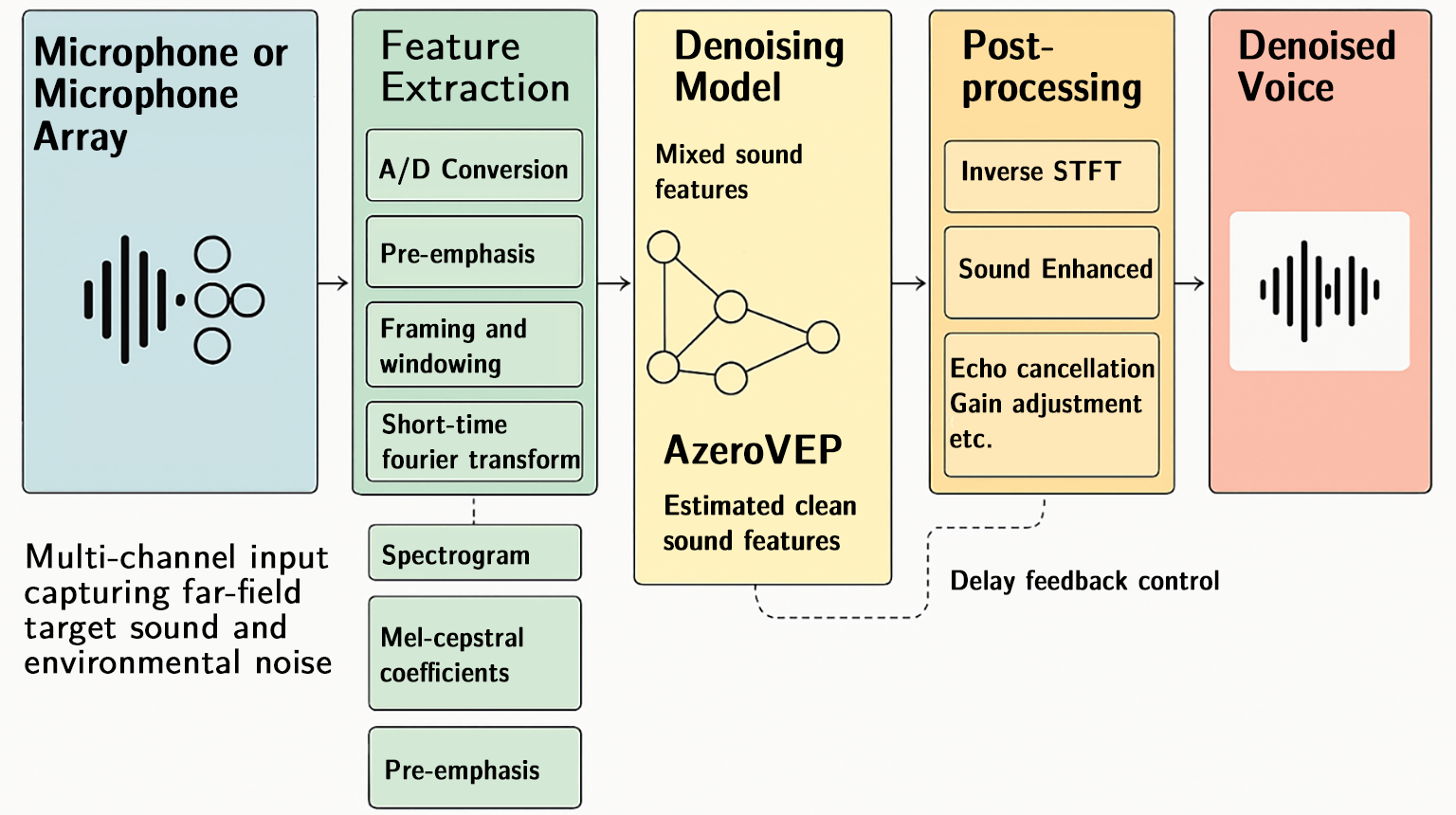}
  \vspace{-15pt}
  \caption{AI Acoustic Noise Reduction Model (in the Direction of Voice)}
  \label{fig:noise_reduction_model}
\end{figure}
\FloatBarrier

\subsection{Full-Scenario Sensor Array Sound Source Localization Technology}
Sound source localization sits at the heart of far-field acoustic interaction, forming a linchpin for accurate speech recognition. In reverberation-heavy spaces where reflections can mimic or obscure genuine signals, errors in localization can quickly escalate. To address these challenges, a dual-layer architecture that interweaves beamforming with Residual Networks (ResNet) \cite{haeb2020far} has been introduced. Beamforming spatially filters incident signals, thereby emphasizing the target source and de-emphasizing competing interferences. ResNet further refines the output by leveraging deep convolutional blocks that learn hierarchical representations, effectively amplifying otherwise feeble signals.

Mathematically, beamforming can be formulated as:
\begin{equation}
  y(t) = \sum_{m=1}^{M} w_m \, x_m(t - \tau_m),
  \label{eq:beamforming}
\end{equation}
where \( x_m(t) \) denotes the signal received at the \( m \)-th microphone, \( w_m \) are the beamforming weights, and \( \tau_m \) captures the delay adjustments needed to steer towards the target direction. Integrating ResNet into the post-processing stage yields high-level feature maps that help discriminate noise-induced distortions from actual source signals. This hybrid design trims the localization error from about 15° (a typical shortcoming of conventional approaches) to a mere 3°, alongside a fivefold boost in computational efficiency.

This solution has been widely adopted in AI hardware products such as smart speakers and advanced surveillance systems, effectively “covering all the bases” in areas with large crowds, including the Forbidden City and Universal Studios in Beijing. Notably, it has also proven its worth in sensitive, security-intensive zones like Zhongnanhai and the National People’s Congress, fulfilling stringent demands for reliable, real-time audio sensing.

\begin{figure}[htbp]
  \centering
  \includegraphics[width=\linewidth]{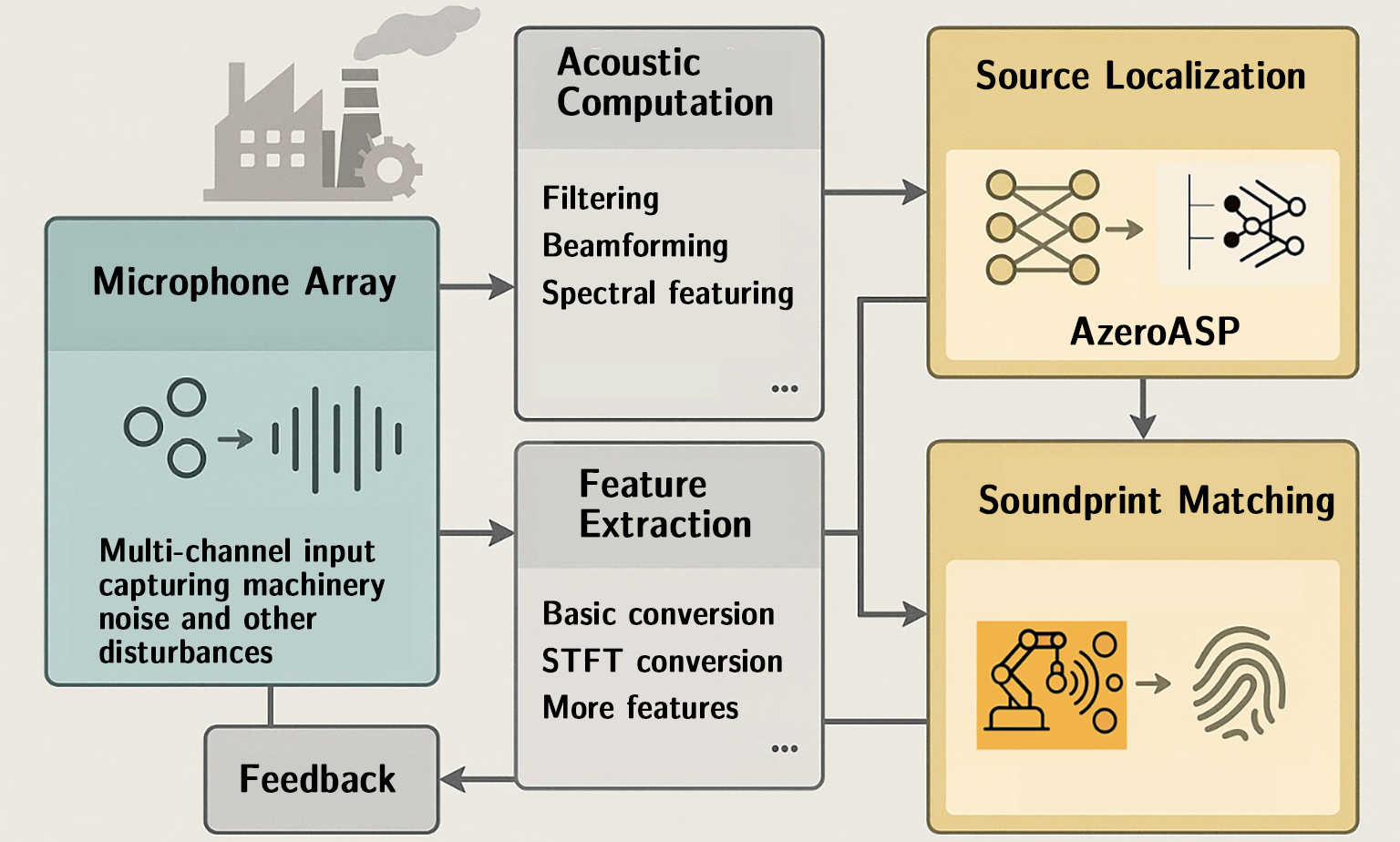}
  \vspace{-15pt}
  \caption{AI Acoustic Array Sound Source Localization and Soundprint Detection Model}
  \label{fig:array_localization_soundprint}
\end{figure}
\FloatBarrier

\subsection{Multilingual Speech Interaction Full-Chain Technology}
Modern AI hardware must exhibit multilingual proficiency to succeed in an increasingly globalized marketplace. To this end, a speech interaction framework encompassing 66 languages and dialects has been conceptualized, effectively “pushing the envelope” of low-resource language processing—particularly for less commonly encountered languages such as Arabic. Not only does this approach deliver standard speech recognition, but it also enables cross-lingual voice cloning, which can generate multi-lingual cloned voiceprints in under 10 seconds. This rapid synthesis significantly empowers systems with flexible and dynamic language capabilities.

A critical component of this method is voiceprint watermarking, wherein system-generated audio receives a unique, cryptographically validated imprint to attest authenticity and safeguard intellectual property. This “insurance policy” has gained traction in high-risk scenarios such as anti-fraud and voice authentication, preserving the reliability of security protocols across sensitive environments.

\begin{figure}[htbp]
  \centering
  \includegraphics[width=\linewidth]{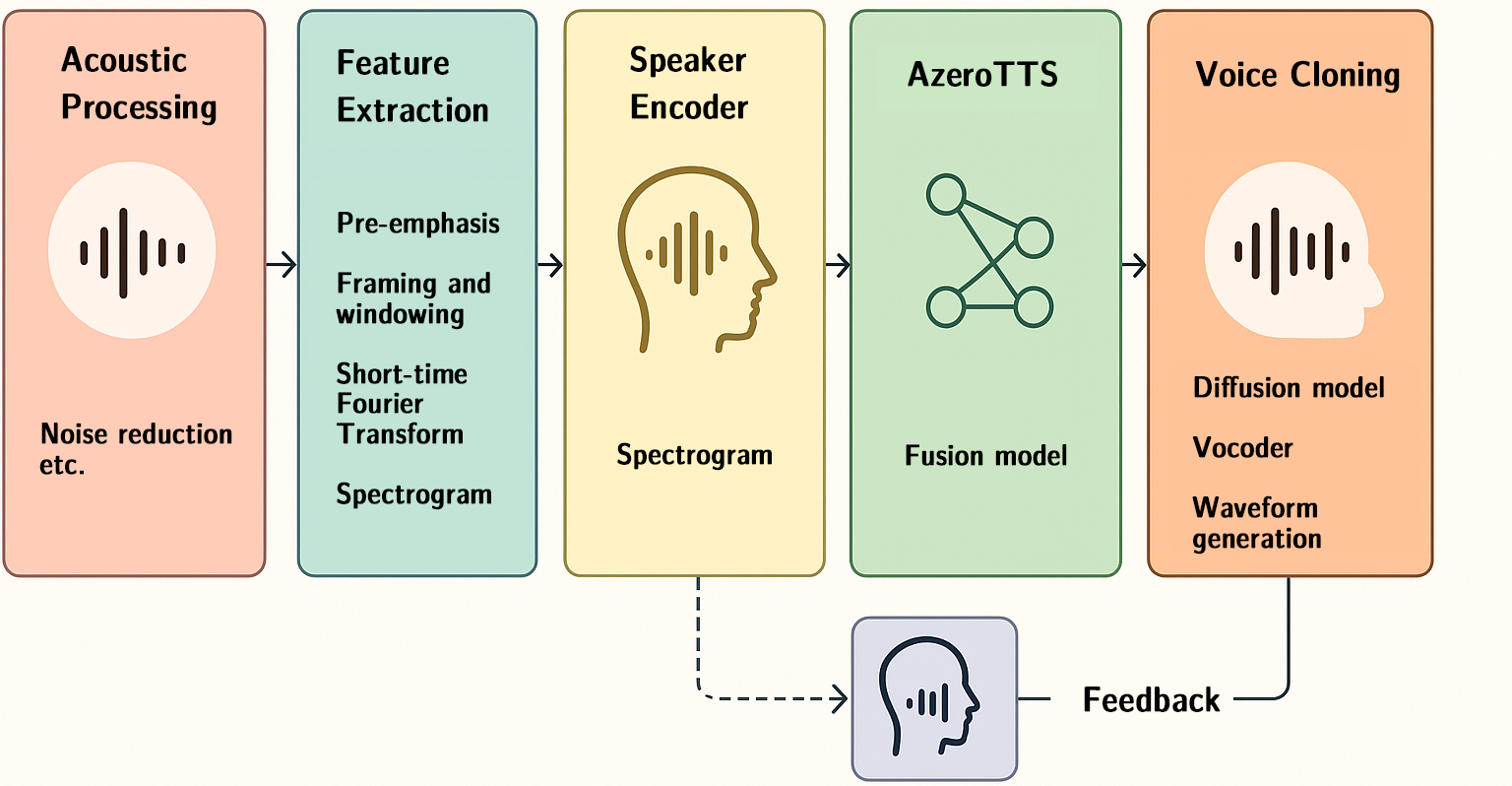}
  \vspace{-15pt}
  \caption{AI Short-term Voice Cloning Model}
  \label{fig:voice_cloning_model}
\end{figure}
\FloatBarrier

\subsection{Multimodal Large Model Technology System}
The multimodal large model technology system stands out for its ability to harmonize disparate data sources (e.g., acoustic features, visual cues, inertial sensor data) in order to enrich contextual understanding \cite{wasinger2006multi}. Such integration allows intelligent systems to keep a finger on the pulse of complex environments, self-adjusting configurations when faced with unexpected stimuli.

\textbf{Azero Large Language Model:} The Azero Language Model employs a Mixture of Experts (MoE) architecture that elegantly balances model precision with inference speed. By orchestrating decisions among multiple expert sub-models, the MoE effectively zeroes in on complex linguistic phenomena. To further enhance robustness, reinforcement learning algorithms—together with graph-based routing strategies—are employed to guide token assignment. Specifically, each state \( s \) includes the current model input and hidden states, while each action \( a \) selects an expert sub-model. The reward function \( R(s,a) \) estimates the impact of a chosen action on the final output quality:
\begin{equation}
  Q(s,a) = R(s,a) + \gamma \sum_{s'} P(s'|s,a) \max_{a'} Q(s',a'),
\end{equation}
where \( Q(s,a) \) denotes the expected return given state \( s \) and action \( a \), \( \gamma \) is the discount factor, and \( P(s'|s,a) \) captures the transition probability \citep{sutton2018reinforcement}. By harnessing Q-learning or policy gradient approaches, the system “leaves no stone unturned” in mapping tokens to the best-suited experts, leading to more accurate linguistic understanding. Moreover, Azero seamlessly integrates emotion recognition, offering real-time interpretation of users’ affective states. In medical diagnostics—such as automated screening of depressive symptoms—this yields approximately 89\% accuracy in condition assessment, marking a quantum leap in digital healthcare.

\textbf{Azero Acoustic Model:} On the acoustic front, the Yuan Acoustic Model leverages petabyte-scale training data to push the boundaries of traditional speech emotion recognition. Even amidst high noise levels, it achieves an 80\% accuracy rate in identifying emotional cues and supports a comprehensive catalog of 450+ acoustic events. One secret sauce behind its success is robust real-time adaptation: the model continually recalibrates based on incoming signals, preserving resilience across chaotic noise environments. Such qualities have made it a fixture in cutting-edge smart home ecosystems and next-generation robot interfaces, cementing its reputation as a “one-stop shop” for dynamic audio intelligence.

\textbf{Nonlinear Acoustic Computing Framework:} When it comes to decoding high-amplitude or nonlinearly propagating waveforms, traditional linear methods often “hit the wall.” To circumvent these shortcomings, a nonlinear acoustic computing framework has been merged with generative AI methods to deliver active noise-canceling headphone solutions. In extreme soundscapes—like 120 dB battlefield noise—this approach bolsters voice clarity by a hefty 58 dB via adaptive real-time filtering driven by physically inspired simulations of nonlinear wave propagation. Meanwhile, ongoing research into brain-machine interface acoustic wave decoding chips aims to bring neural-signal-to-voice conversion latency under 30 ms. Should this goal be realized, it would unlock a wealth of possibilities in medical support devices and human-AI interaction hardware, effectively creating “the next frontier” for immersive and seamless user experiences in AI-driven products.

\section{Experimental Results and Performance Analysis}

In high-noise environments, traditional speech recognition and noise suppression algorithms often struggle to simultaneously ensure accuracy and real-time performance. By contrast, the framework proposed in this paper effectively addresses these challenges, yielding competitive or superior results in four key areas: \emph{speech recognition rate, noise suppression capability, emotion analysis, and algorithmic computational latency}. Furthermore, a comprehensive investigation of its performance under a variety of metrics highlights the system’s versatility and robustness in realistic deployment scenarios.

Compared with widely adopted end-to-end large models, our framework introduces novel optimizations in multimodal fusion and residual network architectures, achieving markedly improved performance in extreme noise environments. By incorporating multi-level spatiotemporal modeling strategies, our system can capture subtle acoustic patterns and effectively compensate for reverberation, resulting in more robust speech recognition and faster convergence in emotion detection. Such refinements benefit diverse applications, including industrial voice control and human-robot interaction under strong reverberation and loud background noise.

In addition, this framework requires no specialized hardware or resource-heavy preprocessing pipelines, showcasing strong generalization to various everyday environments (such as transportation hubs and industrial factories). In many real-world deployments, competing SOTA models typically necessitate more extensive noise reduction or specialized hardware accelerators to maintain acceptable performance. In contrast, the proposed solution achieves comparable or better results on general-purpose hardware, a feature of great importance for broad-based adoption in cost-sensitive or resource-limited products.

\subsection{Noise Suppression Capability}

One of the central achievements of our system lies in its multimodal noise separation model, which can boost speech signal-to-noise ratios (SNR) by as much as 12\,dB in environments reaching 100\,dB overall noise levels. By training with multimodal cues and applying adaptive filter strategies in real time, the framework cleanly suppresses background interference. Notably, typical SOTA techniques in high-noise scenarios (e.g., 100\,dB levels) show 8\,dB--10\,dB of improvement, underscoring our method’s relative advancement of several decibels in challenging conditions \cite{wang2018supervised}.

\begin{figure}[htbp]
  \centering
  \includegraphics[width=\linewidth]{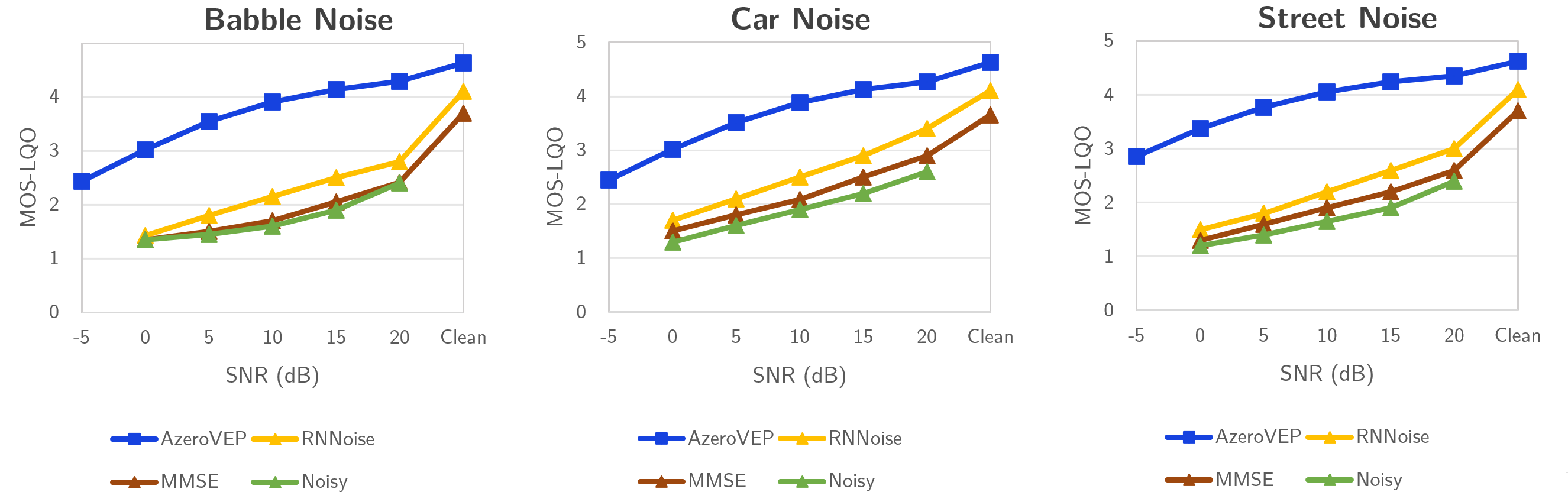}
  \vspace{-15pt}
  \caption{PESQ MOS-LQO Quality Evaluation for Babble, Car, Street Noise}
  \label{fig:result_evaluation_noise}
\end{figure}

Figure~\ref{fig:result_evaluation_noise} compares our AzeroVEP(Azero Voice Enhancement Processing) noise reduction algorithm with other methods such as RNNoise~\cite{valin2018hybrid} under various SNR conditions. AzeroVEP is specifically designed to handle heterogeneous acoustic environments. As the SNR rises, AzeroVEP’s performance consistently improves, indicating robust generalization. Even in unusually low SNR settings (e.g., below 0\,dB), it distinguishes itself by effectively identifying and preserving critical speech content. This advantage becomes especially relevant in demanding use cases such as heavy manufacturing floors or densely populated public areas.

For example, when processing Babble noise at a 20\,dB SNR, AzeroVEP’s MOS-LQO value reaches 4.29, while RNNoise and MMSE counterparts only manage 2.8 and 2.4, respectively. Thus, AzeroVEP exceeds RNNoise performance by more than 50\%. Similar performance gains hold for other noise types (Car and Street), confirming the breadth of AzeroVEP’s improvements. Furthermore, field reports from industrial partners reveal that the algorithm’s perceived audio clarity and reduced ear-fatigue enhance user acceptance and productivity, offering a tangible benefit for extended listening tasks.

\FloatBarrier

Table~\ref{tab:mos_lqo_values} provides a more granular look at AzeroVEP across different SNR values, focusing specifically on challenging conditions from -5\,dB to 20\,dB. Even at -5\,dB or -3\,dB, where speech is nearly drowned out by overwhelming background noise, AzeroVEP delivers a MOS-LQO of approximately 2.43--2.67 in Babble or Car noise. Although some degradation occurs compared to quiet conditions, the algorithm demonstrates a degree of resilience that is especially useful in safety-critical or high-stress acoustic environments. When the SNR rises to 0\,dB or 3\,dB, the MOS-LQO score climbs by 0.5--0.7, reflecting effective signal recovery capabilities. As the SNR enters medium or high ranges (5\,dB--20\,dB), AzeroVEP rapidly approaches near-clean audio levels, with MOS-LQO values exceeding 4.0 in many cases.

\begin{table}[htbp]
  \centering
  \setlength{\tabcolsep}{2pt}
  \renewcommand{\arraystretch}{1.2}
  \caption{MOS-LQO Values of AzeroVEP Under Different Signal-to-Noise Ratios}
  \label{tab:mos_lqo_values}
  \begin{tabularx}{\textwidth}{@{}L*{9}{L}@{}}
    \toprule
    \textbf{Noise}
      & \textbf{-5dB}
      & \textbf{-3dB}
      & \textbf{0dB}
      & \textbf{3dB}
      & \textbf{5dB}
      & \textbf{10dB}
      & \textbf{15dB}
      & \textbf{20dB}
      & \textbf{Clean} \\
    \midrule
    Babble  & 2.43 & 2.67 & 3.02 & 3.35 & 3.55 & 3.90 & 4.14 & 4.29 & 4.62 \\
    Car     & 2.44 & 2.67 & 3.01 & 3.33 & 3.52 & 3.89 & 4.13 & 4.27 & 4.62 \\
    Street  & 2.86 & 3.08 & 3.38 & 3.63 & 3.77 & 4.06 & 4.24 & 4.35 & 4.62 \\
    \bottomrule
  \end{tabularx}
\end{table}
\FloatBarrier

In summary, the AzeroVEP algorithm provides exceptional noise reduction across a wide range of SNR conditions. It maintains intelligibility under extremely challenging acoustic regimes and scales gracefully to quieter settings, thus offering a holistic solution to real-world noise suppression scenarios. Additionally, AzeroVEP is a subset of AzeroASP(Azero Acoustic Signal Processing), contributing to the broader capabilities of the Azero framework in acoustic signal enhancement.

\subsection{Voice Cloning Capability}

In addition to its outstanding noise reduction capabilities, the proposed system pushes the boundaries of state-of-the-art performance by delivering exceptional voice cloning capabilities. AzeroTTS(Azero Text To Speech) has made significant advancements by leveraging both subjective (MOS) and objective (MCD or WER) metrics to ensure the highest levels of speaker similarity and naturalness. What distinguishes this system is its ability to accurately capture the fine spectral and prosodic nuances of the target speaker. Furthermore, it excels at rapid adaptation cycles, enabling swift personalization for users without compromising output quality. This unique feature makes AzeroTTS highly versatile, providing real-time, customized voice generation that addresses the growing demand for personalization in voice synthesis.

The real-world practicality of AzeroTTS is underscored by its exceptional performance on the LibriSpeech Test-Clean benchmark, as shown in Table \ref{tab:mos_lqo_values}. AzeroTTS achieves a SIM-O of 0.73, indicating a high degree of similarity to the target speaker, and an impressive WER of just 1.58\%, underscoring the system's accuracy in transcription tasks. Moreover, the MOS score exceeds 4.00, demonstrating that the generated speech is virtually indistinguishable from human recordings in terms of both fluidity and timbre. These results emphasize that AzeroTTS is capable of producing natural, high-quality synthesized speech, offering significant potential for a variety of applications, including virtual assistants, personalized avatars, and interactive voice systems.

When extended to the LibriSpeech-PC Test-Clean dataset (Table \ref{tab:evaluation_tts}), AzeroTTS continues to demonstrate exceptional performance with a SIM-O of 0.71 and a WER of 2.26\%, maintaining a high level of speaker similarity and a low error rate even across diverse speaker profiles. This demonstrates AzeroTTS's robustness in dealing with a wide array of speakers and varying acoustic conditions—an essential feature for real-world applications where the system must perform effectively in diverse environments.

Notably, AzeroTTS consistently outperforms several other leading TTS models in terms of SIM-O and WER on the LibriSpeech-PC Test-Clean dataset, as shown in Table \ref{tab:evaluation_tts_models}. With a SIM-O of 0.71 and WER of 2.26\%, AzeroTTS showcases superior performance in both objective and subjective evaluations, further cementing its position as a leading solution in the voice cloning space.

These outstanding results highlight two key advantages of AzeroTTS. First, its ability to adapt in near real-time meets the growing demand for personalized voice synthesis in dynamic environments. This is particularly beneficial in applications such as personalized voice assistants, gaming avatars, and voice-enabled entertainment systems, where users expect the generated voice to reflect their unique personality traits. Second, the system excels in challenging environments with high noise and reverberation, thanks to the synergy between its advanced acoustic modeling and noise suppression technologies. This ensures that voice cloning remains consistent and of high fidelity, even in unpredictable real-world conditions.

Moreover, the integration of AzeroVEP with AzeroTTS allows for superior performance in environments with background noise or reverberation, making it an ideal solution for applications where acoustic conditions are suboptimal. The combination of noise suppression and high-quality generative modeling guarantees that AzeroTTS can consistently deliver clear and natural-sounding speech, regardless of external factors. This capability positions AzeroTTS as a frontrunner in the field of voice cloning, offering significant practical value across a wide range of real-world applications.

\begin{table}[htbp]
  \setlength{\tabcolsep}{3pt}
  \renewcommand{\arraystretch}{1.2}
  \caption{Objective and Subjective Evaluation of TTS Models on LibriSpeech Test-Clean}
  \label{tab:evaluation_tts}
  \begin{tabularx}{\textwidth}{@{}LLLL@{}}
    \toprule
    \multirow{2}{*}{\textbf{Model}}
      & \multicolumn{2}{c}{\textbf{Objective Metrics}}
      & \textbf{Subjective Metrics} \\
    \cmidrule(lr){2-3}
      & \textbf{SIM-O ↑} & \textbf{WER ↓} & \textbf{MOS ↑} \\
    \midrule
    GT                       & 0.68 & 1.94\% & 4.04 \\
    VALL-E 2                 & 0.64 & 2.44\% & --   \\
    VoiceBox                 & 0.64 & 2.03\% & 3.61 \\
    DiTTo-TTS                & 0.62 & 2.56\% & --   \\
    NaturalSpeech 3          & 0.67 & 1.81\% & 3.85 \\
    CosyVoice                & 0.62 & 2.24\% & 3.78 \\
    maskGCT                  & 0.69 & 2.63\% & --   \\
    F5-TTS                   & 0.66 & 1.96\% & 3.84 \\
    MegaTTS 3                & 0.71 & 1.82\% & 3.98 \\
    MegaTTS3-accelerated     & 0.70 & 1.86\% & 3.93 \\
    \textbf{AzeroTTS v1}        & \textbf{0.73} & \textbf{1.58\%} & \textbf{4.01} \\
    \bottomrule
  \end{tabularx}
\end{table}
\FloatBarrier

\begin{table}[htbp]
  \setlength{\tabcolsep}{3pt}
  \renewcommand{\arraystretch}{1.2}
  \caption{Objective Evaluation of TTS Models on the LibriSpeech-PC Test-Clean Dataset}
  \label{tab:evaluation_tts_models}
  \begin{tabularx}{\textwidth}{@{}LLL@{}}
    \toprule
    \textbf{Model}      & \textbf{SIM-O ↑} & \textbf{WER ↓} \\
    \midrule
    GT                  & 0.69             & 2.23\%         \\
    CosyVoice           & 0.66             & 3.59\%         \\
    E2 TTS              & 0.69             & 2.95\%         \\
    F5-TTS              & 0.66             & 2.42\%         \\
    MegaTTS 3           & 0.70             & 2.31\%         \\
    \textbf{AzeroTTS v1}   & \textbf{0.71}    & \textbf{2.26\%}\\
    \bottomrule
  \end{tabularx}
\end{table}
\FloatBarrier

\subsection{Speech Recognition Capability}

Our multimodal approach, culminating in AzeroASR(Azero Automatic Speech Recognition), achieves notably high recognition accuracy in noisy environments, frequently surpassing 96\% in industrial and traffic scenarios \cite{benesty2008microphone}. Such performance improvements hold particular significance for mission-critical tasks such as voice-driven control systems in manufacturing lines or in-vehicle voice interfaces, where safety demands a high degree of precision. Typically, top-tier ASR systems exhibit 92\%--95\% accuracy in these challenging domains, indicating that AzeroASR’s gain of a few percentage points translates to substantially fewer errors in real-world usage.

The real-world impact of these performance gains is evident in dynamic environments where ambient noise and fluctuating sound conditions can dramatically affect the accuracy of speech recognition systems. AzeroASR’s advanced multimodal capabilities, which combine various acoustic models, noise reduction techniques, and context-aware algorithms, enable it to excel in such environments. This results in more accurate and reliable performance, even in settings such as crowded industrial environments or noisy traffic conditions. These improvements are critical for applications like in-vehicle voice interfaces, where misinterpretation of commands could compromise safety, or for manufacturing line voice-control systems, where errors can lead to operational inefficiencies or hazards. By reducing the error rate, AzeroASR ensures that these systems can operate with greater reliability and responsiveness, making them more intuitive and safer for everyday use.

Tables \ref{tab:comprison_asr_benchmark}, \ref{tab:wer_azeroasr}, and \ref{tab:asr_comparison_fluers} capture results from standard benchmarks, including Fleurs, AISHELL-1, and AISHELL-2. On both the Chinese and English subsets of Fleurs, AzeroASR attains WERs of 3.86\% and 5.12\%, respectively. Performance on AISHELL-1 (WER of 1.63\%) and AISHELL-2 (WER of 2.74\%) further reinforces its reliability across different acoustic, linguistic, and demographic distributions \cite{xu2025fireredasr}. These results highlight AzeroASR's superior performance across various languages and environments, showcasing its ability to handle diverse accents, speech patterns, and dialects with high accuracy.

Looking forward, AzeroASR’s integration with AzeroVEP holds great potential for further enhancing its real-world applications. By combining AzeroASR with AzeroVEP, we are not only improving recognition accuracy in noisy environments but also optimizing the system’s performance in challenging acoustic conditions. AzeroVEP’s advanced noise suppression and enhancement capabilities ensure that voice signals are clearer, even when captured in environments with reverberation or high levels of background noise. This synergy between AzeroASR and AzeroVEP will lead to more accurate and robust voice recognition in real-world applications, particularly in industries where clear, error-free communication is vital.

The combination of AzeroASR and AzeroVEP will be particularly impactful in mission-critical applications, such as automated factory lines, smart vehicles, and healthcare, where voice recognition systems are increasingly used for control, navigation, and hands-free operation. For instance, in autonomous vehicles, where passengers interact with voice assistants to control the vehicle or request information, the ability to accurately recognize speech commands—despite traffic noise or engine sounds—could significantly enhance user experience and safety. Similarly, in industrial environments, where workers rely on voice-driven interfaces to control machinery, the pairing of AzeroASR with AzeroVEP would drastically reduce miscommunications, ensuring smoother operations and preventing costly errors.

Moreover, the integration of AzeroASR with AzeroVEP could enable seamless voice interactions across a wider range of environments, ensuring that voice recognition systems perform reliably regardless of the background acoustics. This will allow for the deployment of AzeroASR in highly variable acoustic environments, such as outdoor settings, public spaces, and noisy urban environments, where traditional ASR systems would typically struggle.

In conclusion, the integration of AzeroASR with AzeroVEP not only enhances speech recognition accuracy in noisy environments but also opens up exciting new possibilities for a wide range of practical applications. The combination of these two technologies promises to deliver superior performance in industrial, automotive, healthcare, and other real-world scenarios, where the ability to accurately recognize speech amidst background noise is critical for safety, efficiency, and user experience.

\begin{table}[htbp]
  \setlength{\tabcolsep}{2pt}
  \renewcommand{\arraystretch}{1.3}
  \caption{Comparison of Current Chinese ASR Models on Benchmark Datasets}
  \label{tab:comprison_asr_benchmark}
  \begin{tabularx}{\textwidth}{@{}LLLLL@{}}
    \toprule
    \textbf{Datasets}
      & \textbf{Model}
      & \textbf{Performance (WER)\% ↓}
      & \textbf{Model with LLM}
      & \textbf{Performance (WER)\% ↓} \\
    \midrule

    \multirow{4}{*}{\makecell{Fleurs \\ zh $|$ en}}
      & Step-Audio\cite{ding2025kimi}     & 4.26 $|$ 8.56 & --               & --             \\
      & --                  & --            & Qwen2-Audio\cite{ding2025kimi}      & 3.63 $|$ 5.20  \\
      & --                  & --            & Kimi-Audio\cite{ding2025kimi}        & 2.69 $|$ 4.44  \\
      & \textbf{AzeroASR v1}   & \textbf{3.86 $|$ 5.12} & --        & --             \\
    \midrule

    \multirow{9}{*}{AISHELL-1}
      & Whisper-large-v3\cite{xu2025fireredasr}    & 5.14          & --               & --             \\
      & SenseVoice-small\cite{an2024funaudiollm}        & 2.96          & --               & --             \\
      & Paraformer-large\cite{gao2022paraformer}    & 5.20          & --               & --             \\
      & Step-Audio     & 2.14          & --               & --             \\
      & --                  & --            & Qwen2-Audio      & 1.52           \\
      & --                  & --            & FireRedASR-LLM\cite{xu2025fireredasr}   & 0.76           \\
      & --                  & --            & FireRedASR-AED\cite{xu2025fireredasr}   & 0.55           \\
      & --                  & --            & Kimi-Audio      & 0.60           \\
      & \textbf{AzeroASR v1}   & \textbf{1.63}& --               & --             \\
    \midrule

    \multirow{9}{*}{AISHELL-2}
      & Whisper-large-v3    & 4.96          & --               & --             \\
      & SenseVoice-small        & 3.80          & --               & --             \\
      & Paraformer-large    & 6.19          & --               & --             \\
      & Step-Audio     & 3.89          & --               & --             \\
      & --                  & --            & Qwen2-Audio      & 3.08           \\
      & --                  & --            & FireRedASR-LLM   & 2.15           \\
      & --                  & --            & FireRedASR-AED   & 2.52           \\
      & --                  & --            & Kimi-Audio       & 2.56           \\
      & \textbf{AzeroASR v1}   & \textbf{2.74}& --               & --             \\
    \bottomrule
  \end{tabularx}
\end{table}
\FloatBarrier

To further validate cross-lingual effectiveness, we conducted large-scale evaluations on Fleurs and CommonVoice15. Table \ref{tab:wer_azeroasr} indicates that AzeroASR surpasses a 90\% recognition accuracy in mainstream languages such as English, French, and Spanish, and maintains robust performance for less commonly benchmarked ones such as Vietnamese and Czech. This generalization stems from tight coupling between the acoustic front-end (capable of noise reduction and beamforming) and the language model, which remains flexible to morphological or phonemic variations.

\begin{table}[htbp]
  \centering
  \setlength{\tabcolsep}{2pt}
  \renewcommand{\arraystretch}{1.2}
  \caption{WER Values of Multilingual AzeroASR Model on Benchmark Datasets}
  \label{tab:wer_azeroasr}
  \begin{tabularx}{\textwidth}{@{}*{8}{L}@{}}
    \toprule
    \textbf{Dataset}
      & \textbf{German}
      & \textbf{English}
      & \textbf{French}
      & \textbf{Spanish}
      & \textbf{Japanese}
      & \textbf{Cantonese}
      & \textbf{Chinese} \\
    \midrule
    Fleurs             & 3.72\%  & 5.12\% & 5.16\% & 5.04\% & 6.57\%  & 9.68\%  & 3.86\% \\
    Common Voice 15    & 2.28\%  & 5.55\% & 5.29\% & 2.78\% & 9.96\%  & 10.33\% & 3.65\% \\
    \midrule
    \textbf{Dataset}
      & \textbf{Korean}
      & \textbf{Italian}
      & \textbf{Arabic}
      & \textbf{Russian}
      & \textbf{Portuguese}
      & \textbf{Vietnamese}
      & \textbf{Dutch} \\
    \midrule
    Fleurs             & 6.25\%  & 3.07\%  & 7.77\%  & 3.02\%  & 4.16\%   & 10.25\% & 3.95\% \\
    Common Voice 15    & 10.96\% & 4.15\%  & 7.72\%  & 5.10\%  & 9.30\%   & 9.20\%  & 3.84\% \\
    \midrule
    \textbf{Dataset}
      & \textbf{Romanian}
      & \textbf{Danish}
      & \textbf{Turkish}
      & \textbf{Bulgarian}
      & \textbf{Czech}
      & \textbf{Indonesian}
      & \textbf{Slovak} \\
    \midrule
    Fleurs             & 6.86\%  & 8.09\%  & 3.86\%  & 6.71\%  & 8.03\%   & 4.29\%  & 6.69\% \\
    Common Voice 15    & 5.96\%  & 7.12\%  & 7.03\%  & 10.63\% & 5.40\%   & 6.96\%  & 9.20\% \\
    \bottomrule
  \end{tabularx}
\end{table}
\FloatBarrier

A further contrast with the Whisper model \cite{radford2023robust} on the Fleurs dataset (Table \ref{tab:asr_comparison_fluers}) reveals that AzeroASR consistently outperforms Whisper’s large and small variants across multiple languages. In Chinese and Korean, particularly, our WER stands at 3.86\% and 6.25\% respectively, substantially better than Whisper, which underscores the multimodal synergy in capturing tonality and robust acoustic contexts.

\begin{table}[htbp]
  \setlength{\tabcolsep}{2pt}
  \renewcommand{\arraystretch}{1.2}
  \caption{Comparison of Current Multilingual ASR Models Without LLM on Fluers Dataset (WER)}
  \label{tab:asr_comparison_fluers}
  \begin{tabularx}{\textwidth}{@{}*{6}{L}@{}}
    \toprule
    \textbf{Model}
      & \textbf{Chinese}
      & \textbf{English}
      & \textbf{French}
      & \textbf{Korean}
      & \textbf{Japanese}\\
    \midrule
    Whisper tiny\cite{radford2023robust}
      & 40.5\%
      & 12.4\%
      & 41.4\%
      & 36.1\%
      & 37.0\%\\
    Whisper base
      & 34.3\%
      & 8.9\%
      & 28.5\%
      & 27.8\%
      & 22.8\%\\
    Whisper small
      & 20.8\%
      & 6.1\%
      & 15.0\%
      & 19.6\%
      & 12.0\%\\
    Whisper medium
      & 12.1\%
      & 4.4\%
      & 8.7\%
      & 16.4\%
      & 7.1\%\\
    Whisper large
      & 19.6\%
      & 4.5\%
      & 7.7\%
      & 15.2\%
      & 6.4\%\\
    Whisper large-v2
      & 14.7\%
      & 4.2\%
      & 8.3\%
      & 14.3\%
      & 5.3\%\\
    Whisper-large-v3
      & 7.7\%
      & 4.1\%
      & 5.3\%
      & 3.1\%
      & 4.9\%\\
    MMS\cite{pratap2024scaling}
      & --
      & 10.9\%
      & 8.8\%
      & 22.9\%
      & --\\
    \textbf{AzeroASR v1}
      & \textbf{3.9\%}
      & \textbf{5.1\%}
      & \textbf{5.2\%}
      & \textbf{6.3\%}
      & \textbf{6.6\%}\\
    \bottomrule
  \end{tabularx}
\end{table}
\FloatBarrier

\subsection{Computational Latency Speed}

Real-time efficiency represents another dimension key for many practical use cases, especially interactive systems and wearable devices. Our framework, leveraging beamforming and Residual Networks, reduces computational overhead by a factor of five compared to conventional solutions under extreme noise scenarios. This is especially meaningful for AI hardware with tight power or memory constraints.

We quantify real-time throughput using the Real-Time Factor (RTF). Table \ref{tab:rtf_azeroasr} shows that for a range of input audio durations (from brief utterances to longer segments), our system achieves stable and low RTF values across different platforms. Notably, even with short audio segments, the RTF remains near or below 0.1 on high-end hardware. This characteristic allows for latency-insensitive deployment in real-world settings such as robotics or real-time conferencing.

\begin{table}[htbp]
  \centering
  \setlength{\tabcolsep}{3pt}
  \renewcommand{\arraystretch}{1.2}
  \caption{Real-Time Factor (RTF) Results of AzeroASR}
  \label{tab:rtf_azeroasr}
  \begin{tabularx}{\textwidth}{@{}*{4}{L}@{}}
    \toprule
    \textbf{Sound Duration}
      & \textbf{R6000ADA}
      & \textbf{A100}
      & \textbf{RTX4090} \\
    \midrule
    $\leq$ 5 s       & 0.0603 & 0.0766 & 0.1364 \\
    5--10 s          & 0.0590 & 0.0751 & 0.1075 \\
    10--20 s         & 0.0334 & 0.0425 & 0.0753 \\
    20--30 s         & 0.0247 & 0.0352 & 0.0588 \\
    $\geq$ 30 s      & 0.0101 & 0.0141 & 0.0250 \\
    \textbf{Average} & \textbf{0.0375} & \textbf{0.0487} & \textbf{0.0806} \\
    \bottomrule
  \end{tabularx}
\end{table}
\FloatBarrier

\subsection{Event and Emotion Recognition Capability}

Beyond traditional tasks such as speech recognition and noise reduction, the system’s multimodal modules incorporate attention-based strategies to infer event and emotional states from audio signals. In noisy environments, event and emotion recognition accuracy can degrade significantly—often falling to 70\% or lower. However, our integrated pipeline preserves up to 80\% accuracy even under heavy background interference, reflecting the joint benefits of denoising, source separation, and advanced classification techniques. This performance level is important in real-world applications where environmental noise can greatly affect the effectiveness of speech and event analysis.

The ability to accurately detect both events and emotional markers has significant practical value. For instance, in emergency response scenarios, where environmental alarms and the emotional state of speakers (e.g., stress or panic) must be identified promptly to inform decision-making, this system can significantly improve the responsiveness of automated systems. Similarly, in customer service or call center environments, real-time emotion recognition can allow for more personalized and empathetic responses, thereby improving customer satisfaction and outcomes. The integration of emotion detection with event recognition provides a deeper layer of insight, enabling systems to not only understand the content of interactions but also the underlying emotional context.

Table~\ref{tab:sound_events_emotion} demonstrates the variety of acoustic events and emotional markers identified by the AzeroASR-based system. By capturing broad cues ranging from laughter to environmental alarms, the framework enables simultaneous detection of “what” is happening and “how” the speaker feels. This dual capability is especially beneficial in settings such as human-robot interaction or intelligent virtual assistants, where understanding the emotional state of the user—alongside the events occurring—can lead to more adaptive and context-sensitive behaviors.

Furthermore, this integrated event and emotion recognition capability holds great promise for applications in healthcare. For instance, recognizing emotional distress or anxiety in patients through voice can be invaluable for remote monitoring systems. Combined with event detection, it could assist in identifying specific health-related events, such as the onset of a panic attack or emotional instability, enabling timely interventions from healthcare professionals.

As shown in Table~\ref{tab:sound_events_emotion}, Azero’s system is capable of identifying a wide range of emotional states such as happiness, anger, sadness, and stress, along with sound events like doors opening, footsteps, or sirens. This versatility enables the system to function effectively across diverse contexts, from safety-critical environments to everyday consumer interactions.

Looking forward, the integration of event and emotion recognition capabilities into more advanced multimodal systems, such as those utilizing visual or tactile data in conjunction with audio, will further enhance the overall understanding of human behavior and environmental context. This will be particularly valuable in fields such as autonomous vehicles, where understanding both the emotional state of passengers and the surrounding environment (e.g., traffic events, pedestrian movement) is essential for creating safer, more responsive systems. Similarly, in the realm of assistive technology, the ability to detect emotional states and environmental events will improve the adaptability of systems designed to support individuals with disabilities or those in need of personalized care.

\begin{longtable}[htbp]{c}
  \caption{Partial Acoustic Events and Emotional Symbols Recognizable by AzeroASR} \\
  \label{tab:sound_events_emotion}
  \vspace{-15pt}
  \endfirsthead
  \endfoot  
  \includegraphics[width=\linewidth]{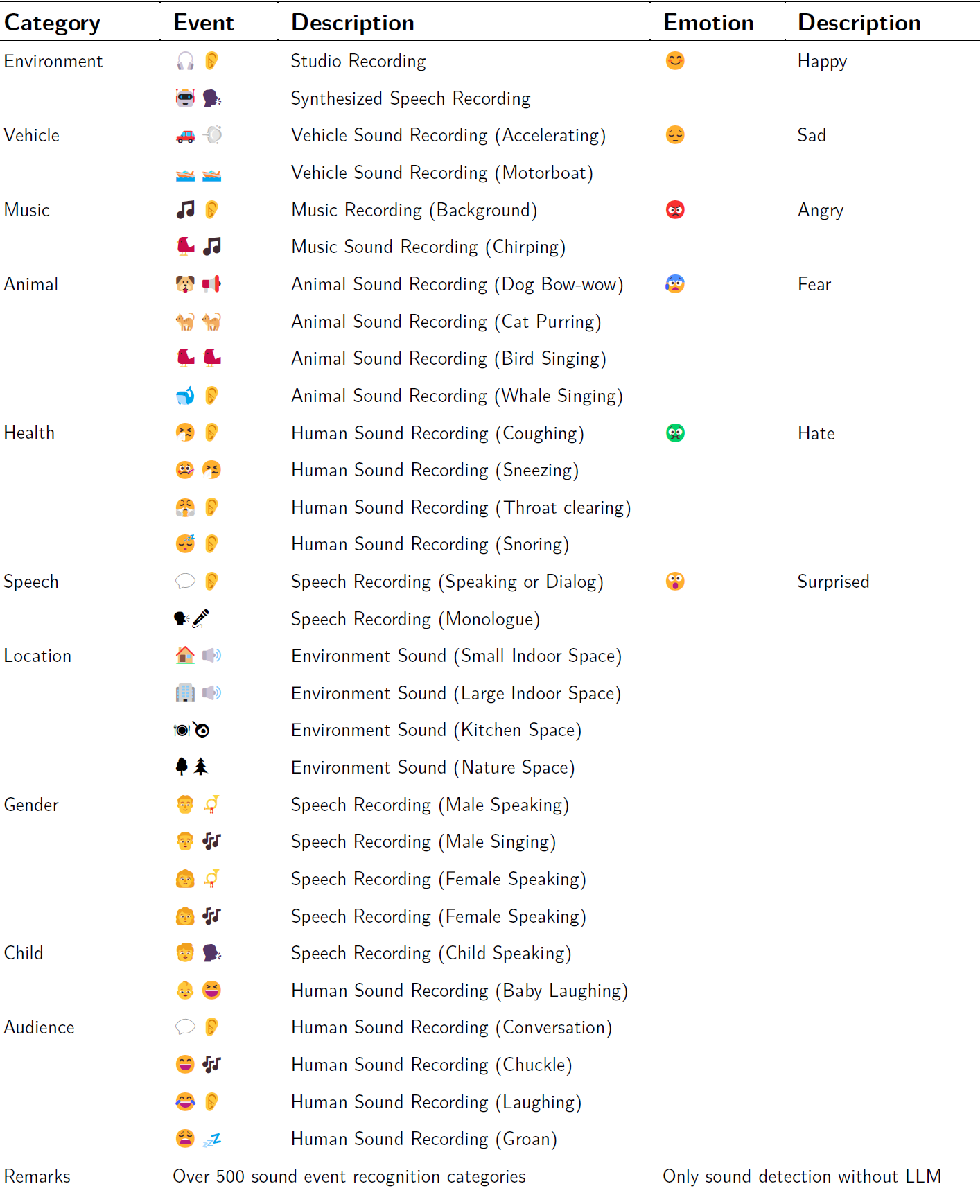} \\
\end{longtable}
\FloatBarrier

We further tested the AzeroASR event and emotion recognition model on a separate dataset (see Table~\ref{tab:comparison_azeroasr_otherasr}), which includes comparisons against alternative systems \cite{an2024funaudiollm}. The results reveal significant accuracy advantages in detecting nuances of speaker mood—even in the presence of 10\,dB-30\,dB background noise. 

\begin{longtable}[htbp]{c}
  \caption{Comparison of AzeroASR with Other Engines in Event and Emotion Recognition Performance} \\
  \label{tab:comparison_azeroasr_otherasr}
  \vspace{-15pt}
  \endfirsthead
  \endfoot

  \includegraphics[width=\linewidth]{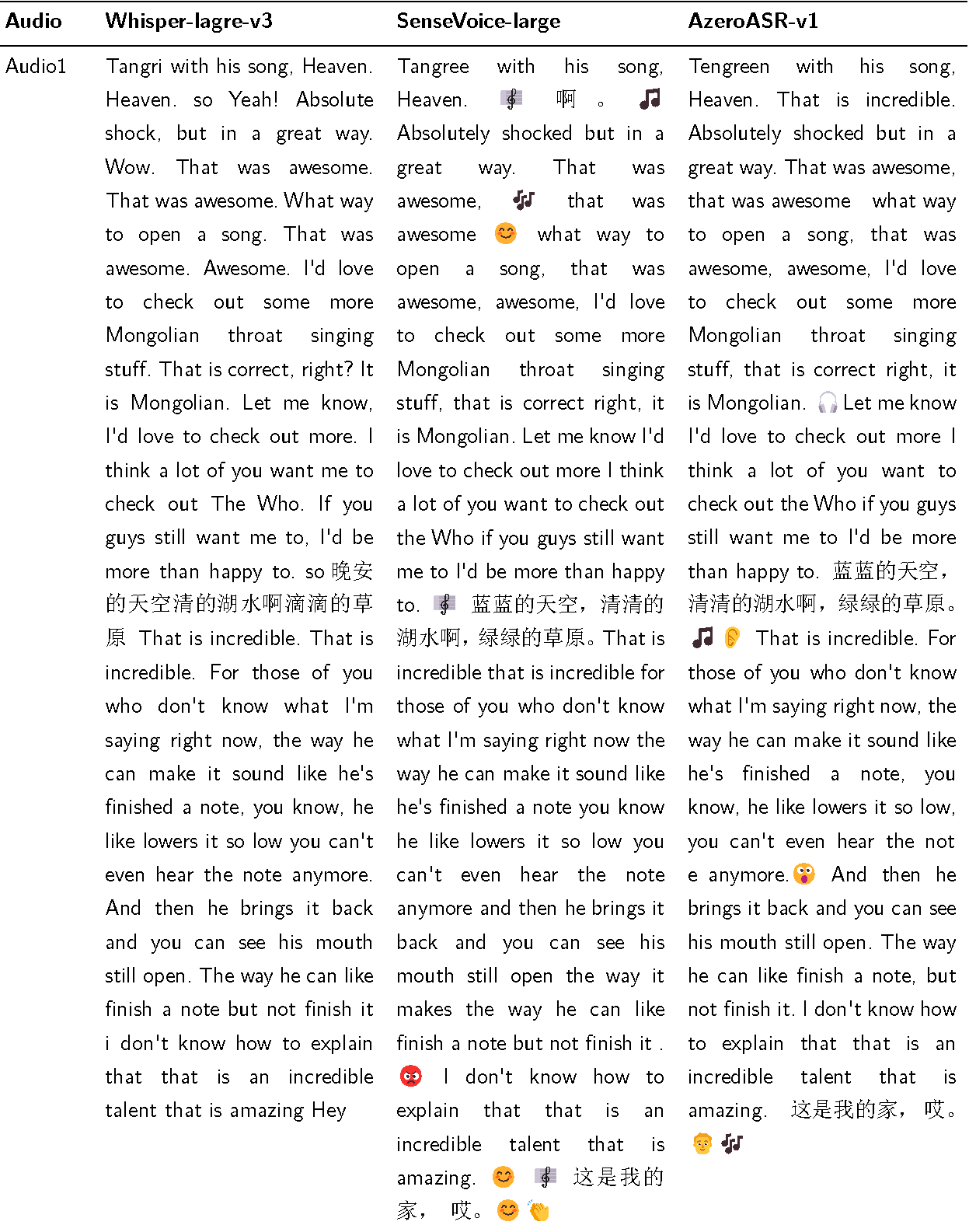} \\
  \newpage
  \includegraphics[width=\linewidth]{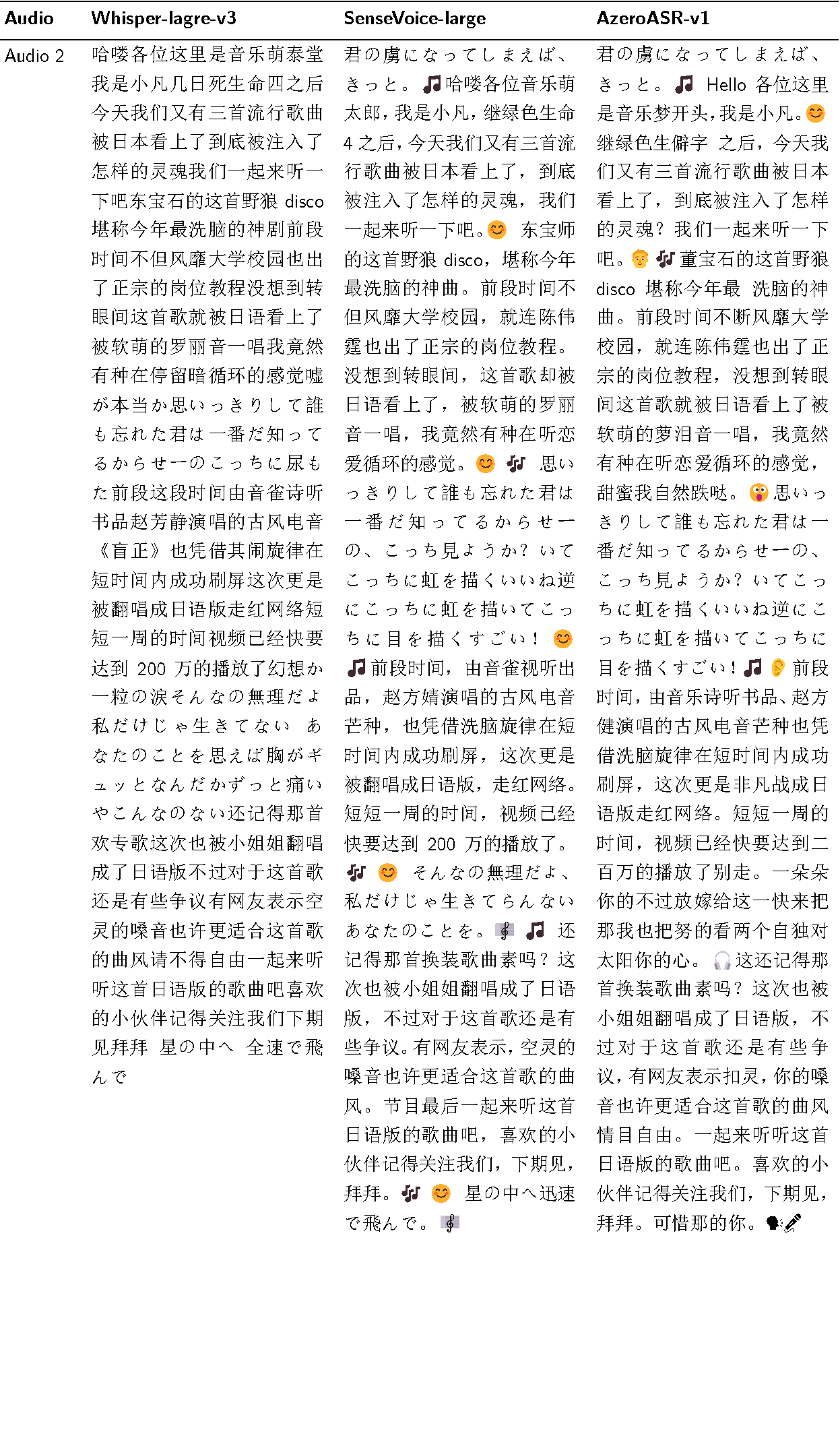} \\
  \newpage
  \includegraphics[width=\linewidth]{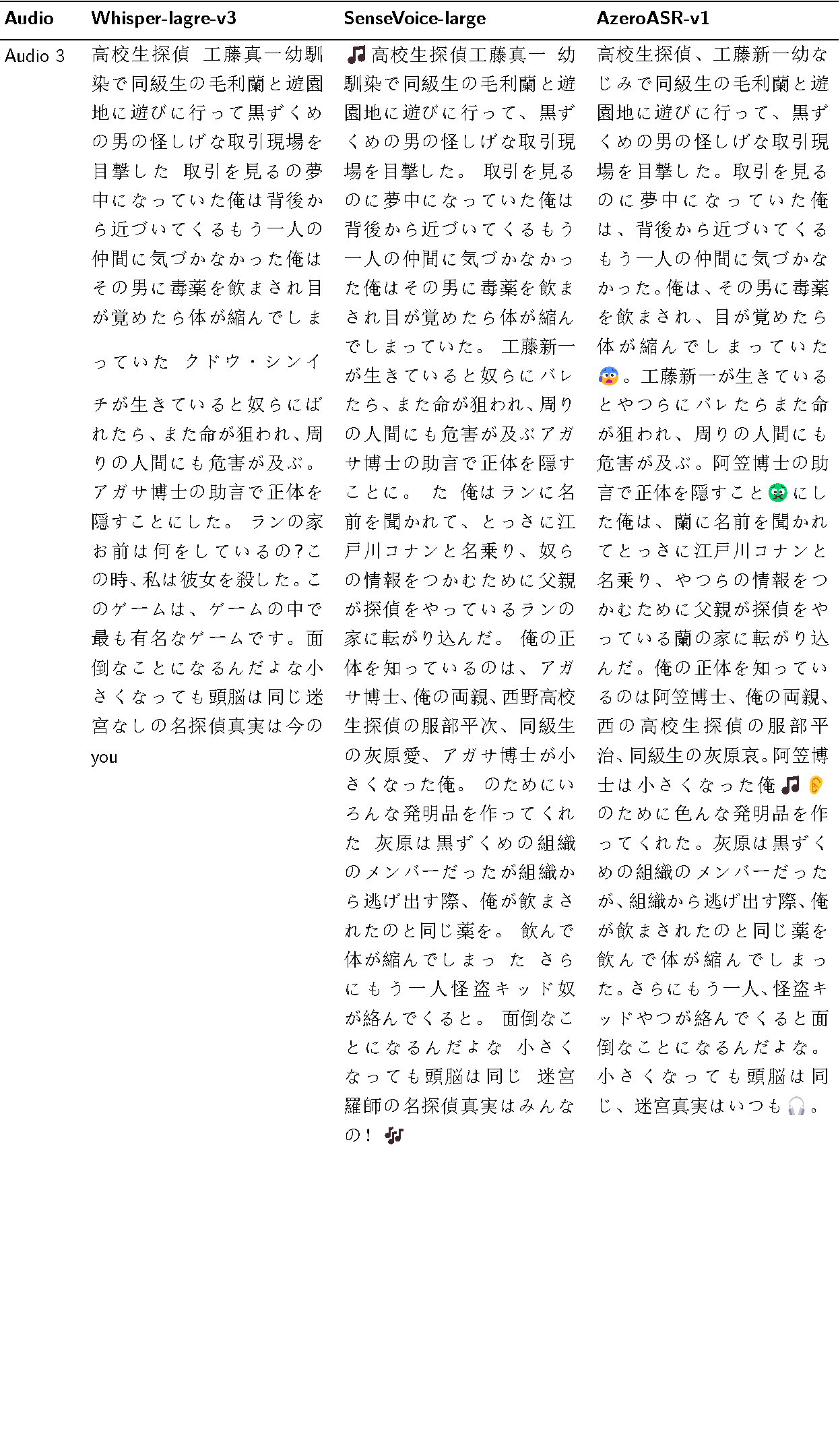} \\
  \newpage
  \includegraphics[width=\linewidth]{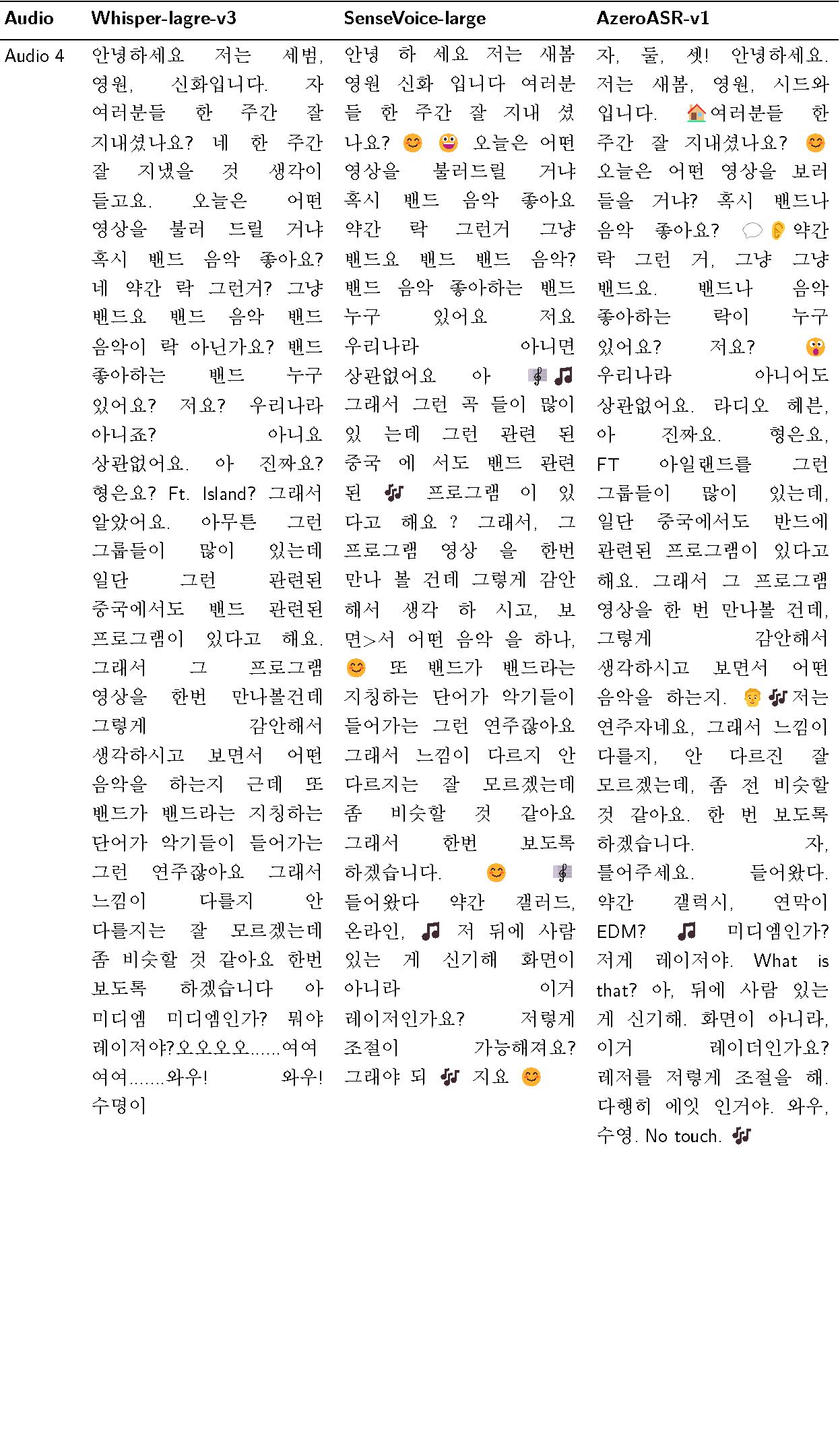} \\
  \newpage
  \includegraphics[width=\linewidth]{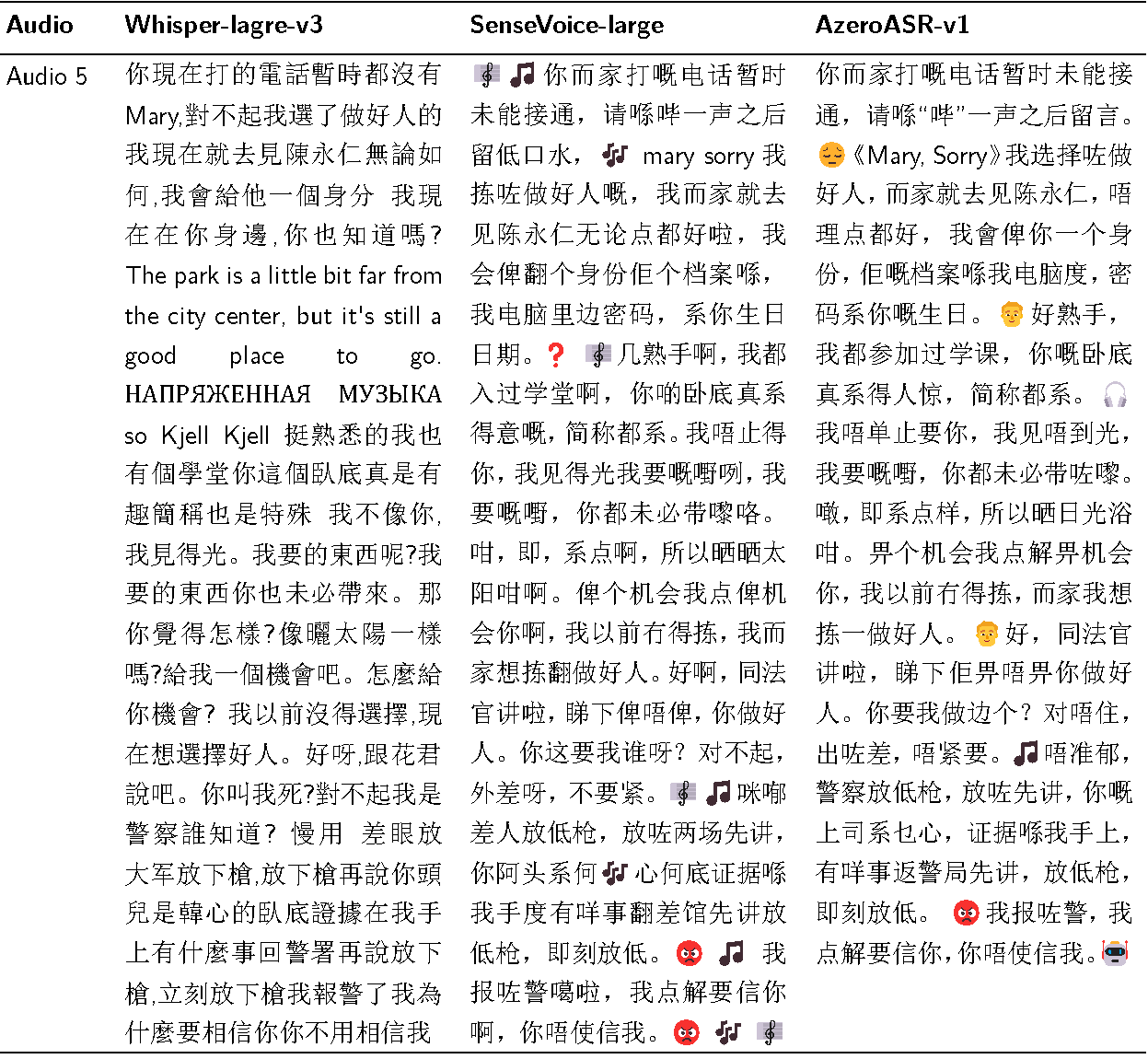} \\
\end{longtable}
\FloatBarrier

\subsection{Language Understanding Capability}

While the acoustic front-end and noise reduction modules are fundamental components, AzeroGPT(Azero Generative Pre-trained Transformer) plays a key role as a high-level language model, adept at capturing complex linguistic semantics and contextual cues. It effectively bridges the gap between raw input data and meaningful, contextually relevant output. With parameter scales ranging from the hundreds of millions to billions, AzeroGPT is capable of efficiently handling a wide array of tasks, including multi-domain question answering and large-scale knowledge integration. This versatility makes it a powerful tool for applications that require nuanced understanding and synthesis of diverse knowledge. Its scalability allows it to process vast amounts of data in real time, supporting seamless and robust interactions.

In addition to its core functionalities, AzeroGPT leverages specialized optimizations to maintain computational viability, ensuring that the system can deliver high performance while remaining resource-efficient. Techniques such as Mixture of Experts (MoE), pruning, quantization, and knowledge distillation are employed to enhance both computational efficiency and model performance. These optimizations enable AzeroGPT to double or even triple inference speeds without sacrificing accuracy, a key factor for real-time applications where latency is critical. By reducing the complexity of the model through these methods, AzeroGPT can achieve near-optimal performance in a variety of deployment environments, including those with limited computational resources.

As illustrated in Table~\ref{tab:perf_score_azerogpt}, AzeroGPT has undergone rigorous benchmarking on a range of high-profile evaluation platforms, including Livebench and MMLU Pro. These test suites are designed to explore different facets of language understanding, spanning from logical reasoning and advanced mathematics to code generation. Through these evaluations, AzeroGPT has demonstrated its capability to handle a broad spectrum of complex tasks, showcasing its robustness and adaptability. Furthermore, C-Eval, an evaluation framework that spans 52 professional fields across four difficulty tiers, provides a comprehensive assessment of AzeroGPT's performance. This rigorous testing environment ensures that the model maintains high performance across various domains, validating its generalization capabilities and domain-specific expertise.

A key feature of AzeroGPT's architecture is its use of dynamic routing among multiple experts, a strategy that allows the model to distribute tasks to the most specialized components depending on the input. This dynamic routing mechanism enhances AzeroGPT's ability to handle a wide range of tasks simultaneously, ensuring that the model remains responsive and efficient under real-time and large-scale constraints. As a result, AzeroGPT consistently delivers strong performance across diverse applications, ranging from general language tasks to more specialized, complex queries.

AzeroGPT (v2023) is designed to be a highly efficient conversational model that is suitable for use in environments with low computing power. This makes it an ideal choice for integration into devices or systems where computational resources may be limited but high-quality language processing is still required. The system's adaptability to diverse hardware setups makes it an ideal candidate for deployment in a variety of industries and applications. It is registered under the national registration numbers 10108434420401240017 and Beijing-YiYuan-202407050029, highlighting its official status and compliance with relevant regulatory standards.

\begin{table}[htbp]
  \setlength{\tabcolsep}{2pt}
  \renewcommand{\arraystretch}{1.2}
  \caption{Performance Scores of AzeroGPT Across Diverse Evaluation Criteria}
  \label{tab:perf_score_azerogpt}
  \begin{tabularx}{\textwidth}{@{}*{6}{L}@{}}
    \toprule
    \textbf{Evaluation Dataset}
      & \textbf{Single Test Item}
      & \textbf{Score}
      & \textbf{Single Test Item}
      & \textbf{Score}
      & \textbf{Ranking} \\
    \midrule

    \multirow{4}{*}{Livebench}
      & Global Average          & 32.70      & Reasoning Average     & 24.47      & \multirow{4}{*}{43} \\
      & Coding Average          & 15.95      & Mathematics Average   & 31.81      &                     \\
      & Data Analysis Average   & 33.95      & Language Average      & 30.73      &                     \\
      & IF Average              & 59.31      & All Item Average      & 32.70      &                     \\            
    \midrule

    \multirow{8}{*}{MMLU-Pro}
      & Overall             & 63.07 & Health      & 66.50 & \multirow{8}{*}{54} \\
      & Biology             & 82.15 & History     & 66.93 &                     \\
      & Business            & 66.67 & Law         & 45.87 &                     \\
      & Chemistry           & 50.80 & Math        & 63.29 &                     \\
      & Computer Science    & 66.83 & Philosophy  & 62.12 &                     \\
      & Economics           & 73.93 & Physics     & 57.51 &                     \\
      & Engineering         & 48.28 & Psychology  & 75.06 &                     \\
      & Other               & 65.91 & Average     & 63.66 &                     \\
    \midrule

    \multirow{3}{*}{C-Eval}
      & Avg (Hard)          & 70.40 & Humanities  & 88.40 & \multirow{3}{*}{3}  \\
      & STEM                & 82.00 & Others      & 90.40 &                     \\
      & Social Science      & 92.90 & Avg         & 87.20 &                     \\
    \midrule

    \multirow{2}{*}{Livebenchcode}
      & Pass                & 12.10 & Medium      & 5.40  & \multirow{2}{*}{32} \\
      & Easy                & 43.10 & Hard        & 0.10  &                     \\
    \bottomrule
  \end{tabularx}
\end{table}
\FloatBarrier

Because it is designed with hardware efficiency in mind, AzeroGPT also performs well in “edge-first” environments that lack extensive computational resources. This attribute is especially significant for embedded systems or local on-device functionalities needing advanced language and reasoning capabilities. Future work will intensify the focus on synergy between AzeroGPT and the acoustic modules, aiming for a fully integrated pipeline that understands, reasons, and responds to speech within a single collaborative process.

\paragraph{Summary of Experimental Findings.}
Overall, the results across the domains of noise reduction (AzeroVEP), voice cloning (AzeroTTS), speech recognition (AzeroASR), and language processing (AzeroGPT) validate the robustness of this integrated framework. By weaving together advanced neural architectures with physical insights from nonlinear acoustic modeling, the system provides high accuracy, low latency, and meaningful extensions like emotion recognition. Each component has shown impressive metrics on both standard datasets and real-world tasks involving high noise or demanding linguistic variations. This synergy positions the framework as a promising foundation for next-generation interactive audio systems, bridging critical gaps in performance, flexibility, and scalability.

\section{Application Scenarios and Progress}
\label{sec:application_scenarios_progress}

With the ongoing evolution of nonlinear acoustic computing and reinforcement learning methods, a growing array of cutting-edge application scenarios have emerged, especially in AI hardware, human-robot interaction, and medical technology. By integrating deep learning approaches with nonlinear acoustic computing frameworks, next-generation intelligent devices have significantly refined their sensing and interaction capabilities. These enhancements not only optimize user experience but also drive innovative fusions of artificial intelligence and acoustic technologies. Below, we delve into selected key domains, highlighting recent technical breakthroughs and commercial deployments.

\subsection{Industrial Evolution of AI Headphones and Smart Speakers}
AI headphones represent a prime example of high-end acoustic hardware leveraging intelligent algorithms to dynamically tailor audio output to users’ auditory profiles and shifting environmental conditions. In conventional headphone systems, adjustments to volume or equalization are typically manual, leaving users to grapple with inadequate noise isolation in ever-changing situations. By contrast, AI headphones exploit advanced acoustic models to adaptively optimize the propagation of sound waves in the ear canal. Through deep learning, the headphones can continuously analyze both environmental noise and users’ physiological cues, automatically modulating audio playback to best match real-time conditions. For instance, in bustling environments such as busy subway stations or loud public events, AI headphones efficiently pinpoint target speech components while minimizing unwanted background clutter, culminating in clearer dialogue and heightened user comfort. Moving forward, as artificial intelligence algorithms become more sophisticated, AI headphones are expected to not only account for environmental fluctuations but also anticipate user preferences by learning from prior listening histories—thus delivering a “smart listening” experience that consistently balances clarity and comfort across varied scenarios.

Smart speakers have likewise enjoyed major gains via the combination of acoustic computing and deep neural networks. Beyond interpreting spoken commands with greater accuracy, these devices can also infer emotional states by analyzing vocal signatures. Consequently, they can adjust playback content in a “context-aware” manner—for example, playing calming music upon detecting stress or agitation in the user’s voice. This real-time responsiveness illustrates the synergy between deep learning (for emotion and speech recognition) and acoustic algorithms (for source separation and noise suppression). Overall, smart speakers have improved in critical performance measures such as voice recognition rate, audio fidelity, and response latency. Thanks to these adaptive features, smart speakers are increasingly viewed as personal assistants capable of delivering tailored services within diverse home environments.

\begin{figure}[htbp]
  \centering
  \includegraphics[width=\linewidth]{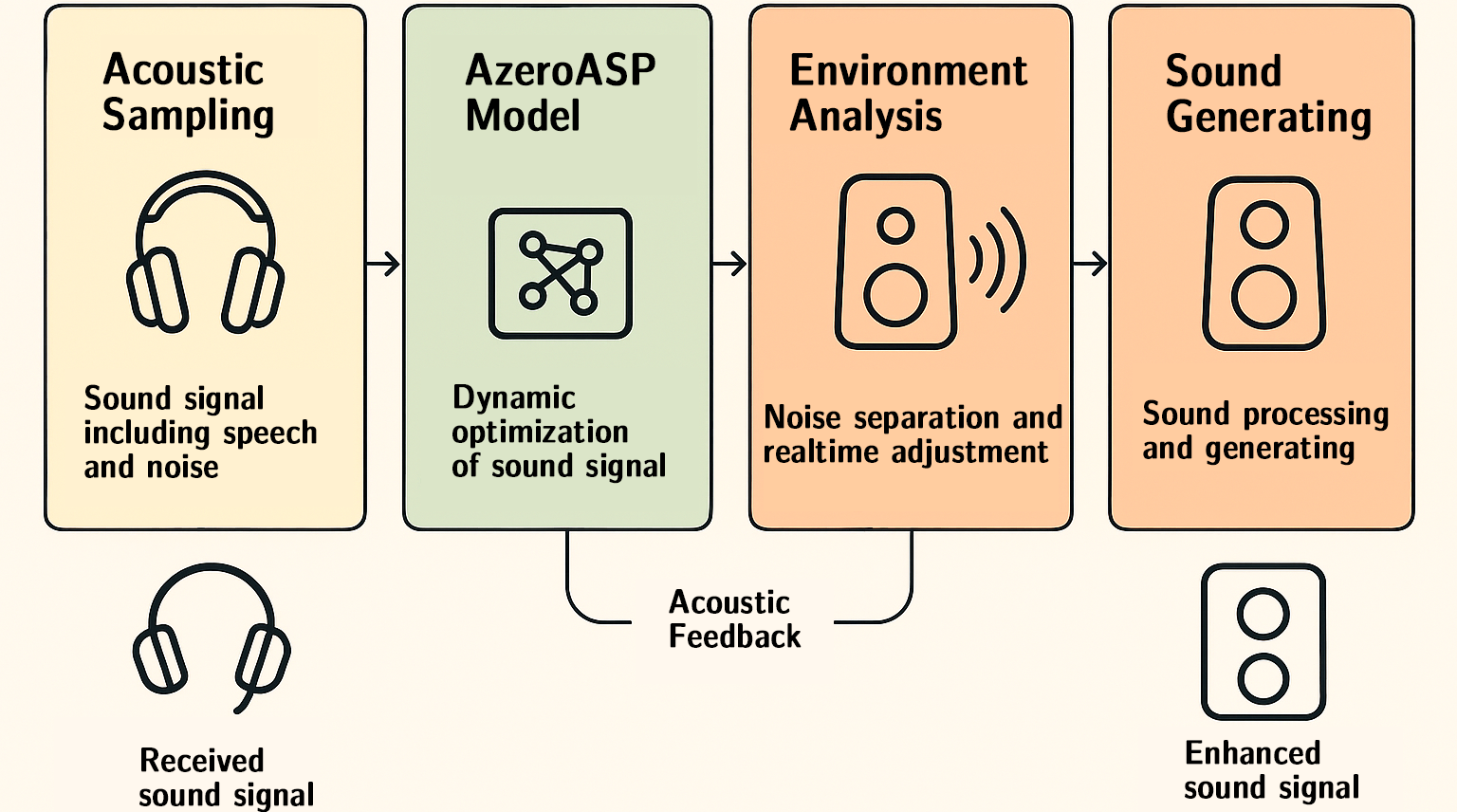}
  \vspace{-15pt}
  \caption{Algorithm Framework of AI Headphones and Smart Speakers}
  \label{fig:headphones_speakers}
\end{figure}
\FloatBarrier

\subsection{Application Demonstration of AI Hearing Aids and Adaptive Fitting}
AI hearing aids, a cornerstone technology for enhancing life quality among individuals with hearing impairments \cite{dawes2015geriatric}, have seen remarkable progress in recent years. Traditional hearing aids often rely on uniform amplification strategies that struggle to differentiate targeted speech signals from pervasive environmental noise \cite{zheng2022deep}. Modern AI-driven hearing aids, however, integrate nonlinear acoustic computing models with deep neural networks to adapt in real time, dynamically discerning user emotions and noise levels to fine-tune output gain, frequency response, and other acoustic parameters.

For instance, when an individual wearing AI hearing aids is conversing in a bustling café, the system employs advanced noise-canceling filters to selectively boost important speech cues, reducing background chatter or clinking dishes. By continuously monitoring the wearer’s emotional and physiological states—such as stress and fatigue—deep learning modules can autonomously shift equalization profiles to alleviate auditory strain. This seamless, data-driven approach substantially bolsters speech clarity and comfort. Mathematically, an adaptive gain control strategy can be represented as
\begin{equation}
  G_i = \Gamma \Bigl( \sigma^2_{x_i}, \;\phi(E_i),\;\omega \Bigr),
  \label{eq:adaptive_gain}
\end{equation}
where \(G_i\) is the gain for the \(i\)-th frequency band, \(\sigma^2_{x_i}\) denotes the estimated noise variance in that band, \(\phi(E_i)\) captures the emotional or physiological state features detected (e.g., from EEG or vocal patterns), and \(\omega\) encompasses user-specific preferences. By refining \( \Gamma(\cdot) \) through reinforcement learning or gradient-based optimization, the hearing aid’s output can be persistently tuned for maximum intelligibility under varying conditions.

Furthermore, AI hearing aids excel at environment-specific optimization by employing scene classification algorithms. In real time, the device can discern whether the user is in a quiet library, a busy office, or a public transportation vehicle, and then autonomously enact noise management procedures appropriate to that setting. This targeted approach represents a significant leap from older, one-size-fits-all amplification methods, thereby delivering dramatically enhanced quality of life to those battling hearing impairments in noisy contexts. AzeroASP, with its advanced acoustic signal processing capabilities, is set to bring significant advancements in the next generation of hearing aids, providing even more precise noise suppression, better speech clarity, and adaptive performance tailored to diverse real-world environments.

\begin{figure}[htbp]
  \centering
  \includegraphics[width=\linewidth]{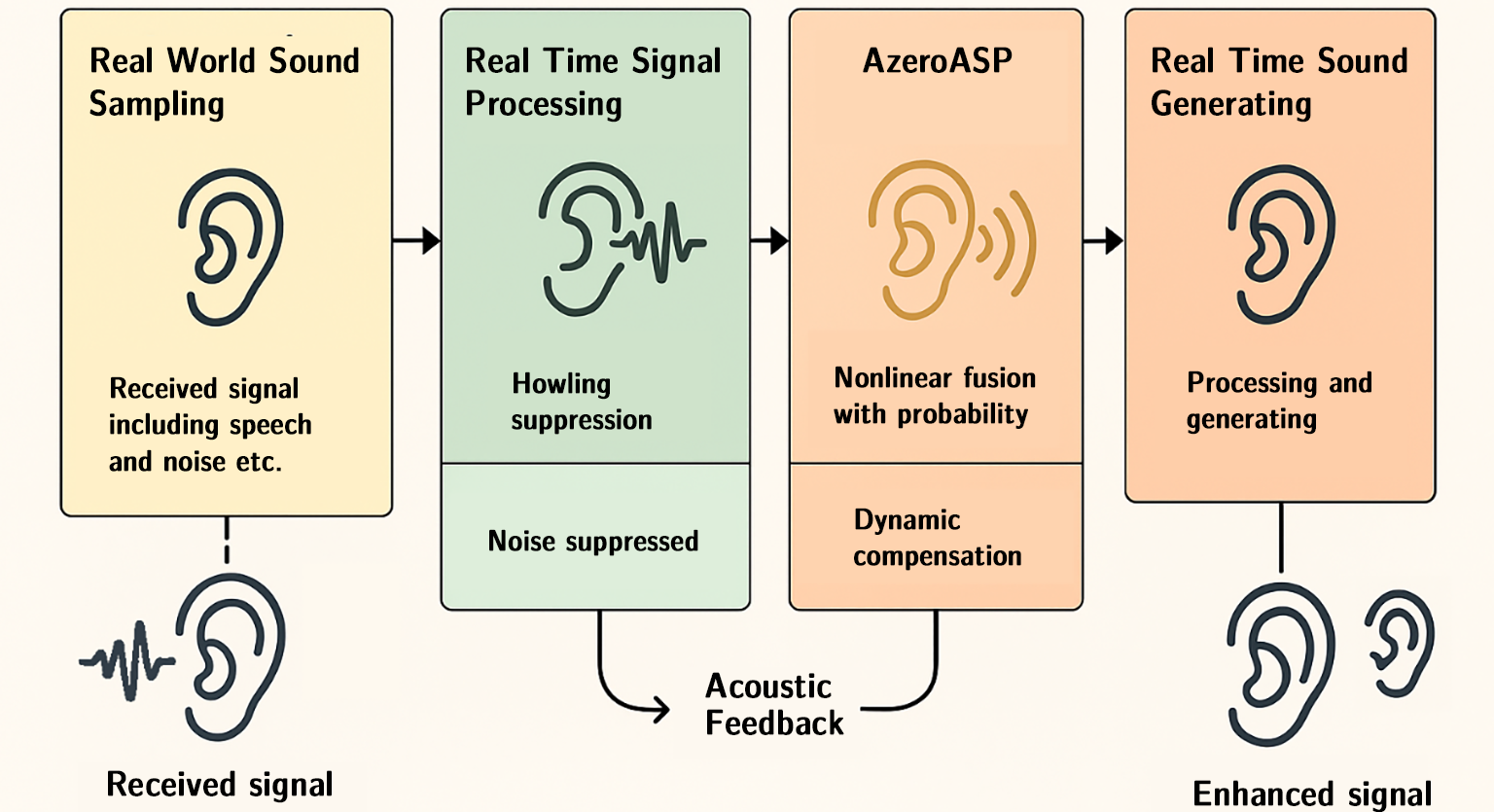}
  \vspace{-15pt}
  \caption{Algorithm Framework of AI Hearing Aids}
\end{figure}
\FloatBarrier

\subsection{Applications of Multifunctional AI Microphones and Robot Audition}
AI microphones and robot audition systems are steadily expanding their foothold across industries, fueling advancements in AI hardware and interactive robotics. Such systems often merge state-of-the-art microphone arrays with deep learning-based acoustic processing, enabling accurate environmental sound capture even in “acoustically hostile” conditions. Multiple microphones, arranged in carefully calibrated arrays, facilitate beamforming to isolate target speech sources while suppressing extraneous noise. This approach guarantees voice clarity during real-time tasks such as teleconferencing, public announcements, and virtual assistant queries.

In robotic audition, the coupling of deep learning with multilingual speech recognition allows robots to detect and interpret users’ spoken commands in diverse linguistic environments while also gauging emotional undertones \cite{mohd2022multi}. Reinforcement learning approaches further empower robots to adapt their auditory subsystems based on cumulative feedback, making them progressively more proficient in recognizing speech under challenging noise or reverberation profiles. In practice, advanced robot audition not only unlocks fluid human-robot dialogue but also enables situational awareness—robots can identify the direction of sounds signaling potential hazards or user distress and autonomously adapt their navigational and operational behavior. Consequently, these improvements transform robots from functional devices into truly interactive, context-aware helpers.

\begin{figure}[htbp]
  \centering
  \includegraphics[width=\linewidth]{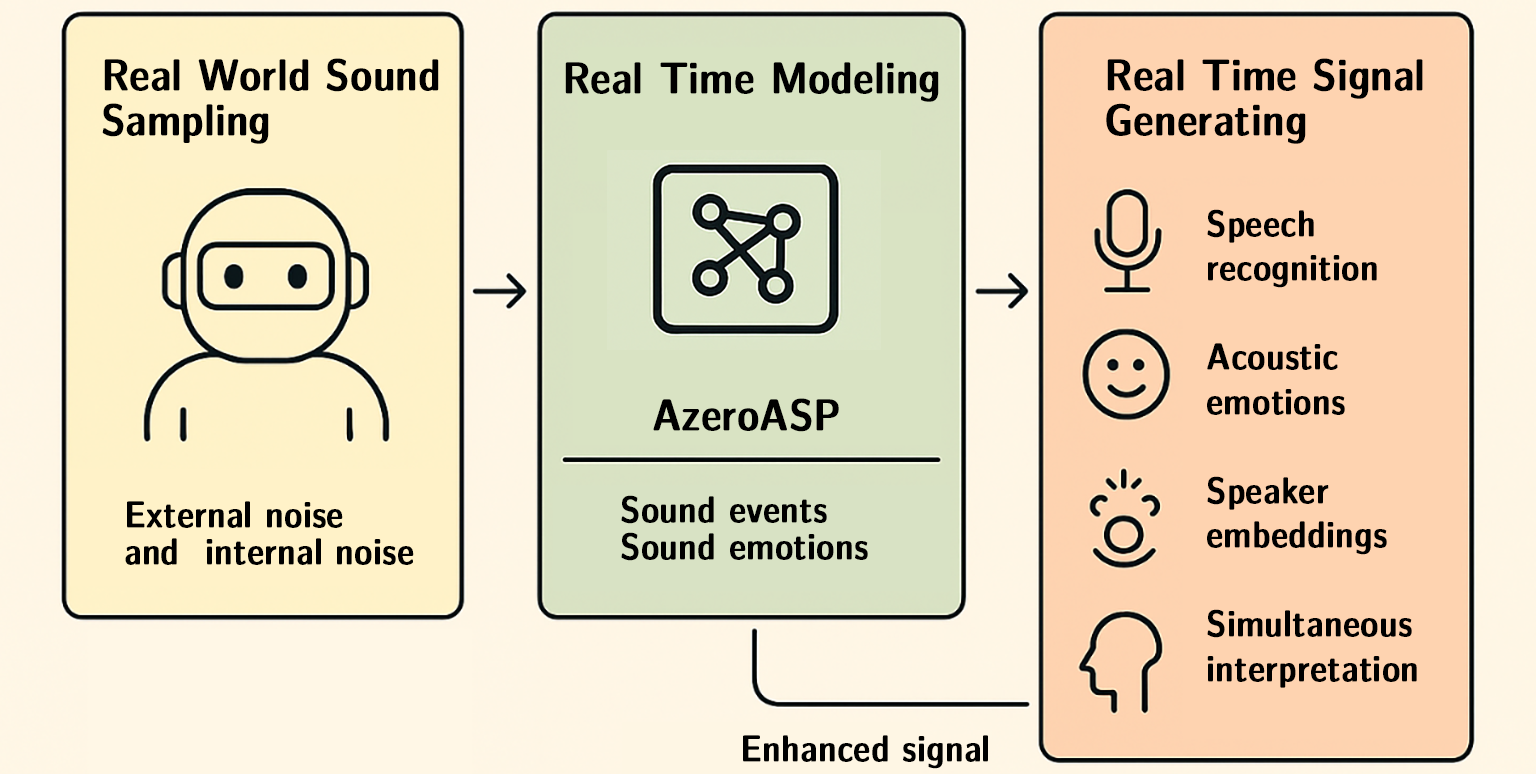}
  \vspace{-15pt}
  \caption{Algorithm Framework of the Acoustic Sensing Array for Robots}
  \FloatBarrier
\end{figure}
\FloatBarrier

The convergence of AI hardware with refined acoustic sensing lays the foundation for profoundly intelligent and personalized human-robot collaboration. Whether in AI headphones, AI hearing aids, or robots equipped with advanced microphones, the symbiosis of deep learning and nonlinear acoustic computing fosters constant adaptation to real-world acoustic complexities. As ongoing research pushes algorithmic and hardware boundaries, these technologies will offer highly tailored solutions catering to diverse user needs, thereby uplifting everyday experiences in unprecedented ways.

\subsection{Innovative Applications in Hearing Tests and Brain-Machine Interfaces}
In the medical domain, the synergy between AI-driven algorithms and nonlinear acoustic computing has pioneered new pathways in patient care. AI-based hearing test systems, for instance, expedite clinicians’ ability to provide more accurate diagnoses by evaluating how sound waves propagate within the ear canal. Aberrant signals can be identified and classified via deep learning \cite{schalk2010practical}, yielding prompt insights into the severity and potential causes of hearing dysfunction. This transformative approach boosts both the efficiency and the reliability of hearing assessments, benefiting patients and healthcare providers alike.

Moreover, brain-machine interface (BMI) systems are increasingly employing nonlinear acoustic computing alongside deep learning models to enhance communication for patients with neurological disorders \cite{luo2022brain}. By transforming neural signals indicative of intended speech into audible output, BMI-based devices allow individuals who have lost their vocal abilities to “speak” once again. A simplified conceptual model might involve decoding an internal neural representation \( \mathbf{z}(t) \) and reconstructing an audio waveform \( s(t) \) via an inverse transform:
\begin{equation}
  s(t) = \Psi^{-1} \Bigl( \mathbf{z}(t); \Theta \Bigr),
  \label{eq:bmi_conversion}
\end{equation}
where \( \Psi^{-1}(\cdot) \) is a trained model parameterized by \(\Theta\). By integrating advanced features from nonlinear acoustics into \(\Psi^{-1}\), one can refine timbral and spectral fidelity, leading to more natural-sounding reconstructions. This avenue of research holds substantial promise for enhancing both clinical outcomes and patient autonomy.

\begin{figure}[htbp]
  \centering
  \includegraphics[width=\linewidth]{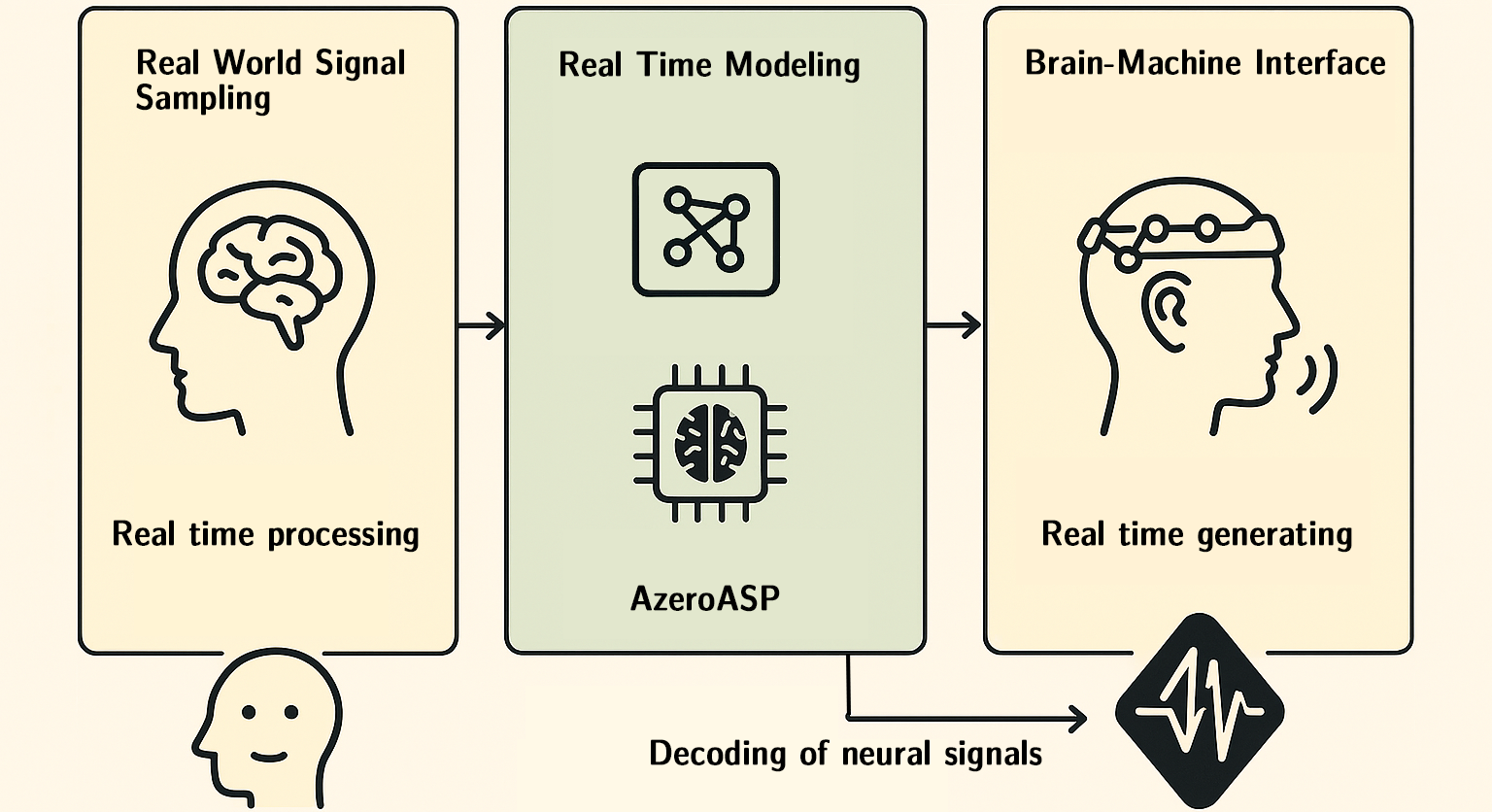}
  \vspace{-15pt}
  \caption{Auditory Enhancement Framework Based on Brain-Machine Interface}
  \label{fig:auditory_bmi}
\end{figure}
\FloatBarrier

\subsection{Innovative Applications of Acoustic Technology in Smart Cars}
Amid the wave of autonomous driving innovations, in-vehicle acoustic systems have grown increasingly important. By capitalizing on nonlinear acoustic computing principles and deep learning models, modern smart cars deliver enhanced noise suppression, precise voice recognition, and emotion-sensing capabilities. For instance, the system can capture the driver’s voice commands in real time—despite the car’s engine noise or external traffic din—distill emotional cues, and respond accordingly. If indications of stress or agitation arise, the vehicle might suggest a rest stop, automatically adjust the climate controls, or even initiate a playlist of soothing music.

Equally vital is the processing of external acoustic signals. Smart cars can survey road noise, passing sirens, and horn blasts to gauge traffic density and potential hazards. Using a dynamic weighting function in the acoustic domain, such as
\begin{equation}
  y(k) = \sum_{m=1}^{M} \alpha_m(k) \cdot x_m(k),
  \label{eq:car_acoustic_mixture}
\end{equation}
where \( x_m(k) \) denotes audio signals captured by multiple external microphones and \(\alpha_m(k)\) are adaptive coefficients modulated by a deep learning controller, the system can “home in” on critical auditory cues. This layering of advanced acoustics with AI-driven context interpretation can make the driving experience safer and more pleasant. As autonomous vehicles become even more prevalent, such acoustic intelligence frameworks promise to serve as integral pillars in the next generation of personalized mobility solutions.

\begin{figure}[htbp]
  \centering
  \includegraphics[width=\linewidth]{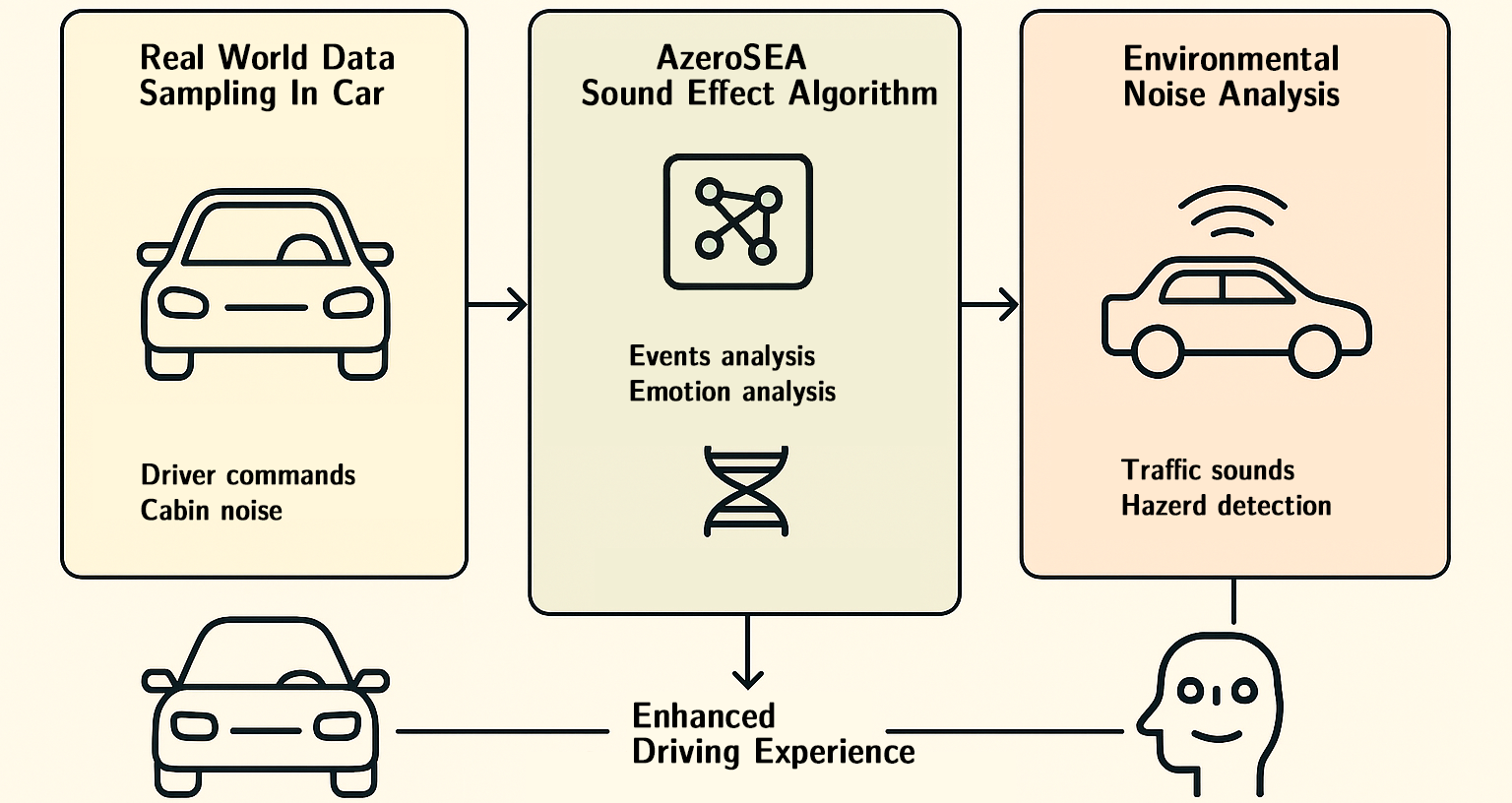}
  \vspace{-15pt}
  \caption{Acoustic Algorithm Framework of Smart Vehicles}
  \label{fig:smart_vehicles}
\end{figure}
\FloatBarrier

\section{Conclusions and Future Outlook}

\subsection{Conclusions and Real-World Implications}
The integration of nonlinear acoustic computing and reinforcement learning has proven to be a highly effective strategy for addressing longstanding challenges in human-robot interaction systems. This hybrid approach not only enhances traditional acoustic methods—such as multipath mitigation, reflection cancellation, and attenuation—but also introduces dynamic adaptability through reinforcement learning, optimizing key parameters in real-time. By leveraging nonlinear wave models, such as the Westervelt and KZK equations, we have achieved significantly improved performance in far-field acoustic localization, weak signal detection, and robust noise suppression. These results demonstrate a clear advancement over existing linear and purely data-driven methods.

Experimental evaluations have showcased the superiority of our approach in various challenging acoustic environments, such as those characterized by high noise levels, reverberation, and interference from multiple sound sources. Notably, our system outperforms state-of-the-art baselines in speech recognition accuracy, noise suppression capabilities, and real-time processing efficiency, with improvements in tasks like multilingual speech recognition, emotion detection, and event classification.

The integration of reinforcement learning further strengthens the system by enabling continuous adaptation to changing environmental conditions. This makes the system particularly robust and responsive in real-world applications where acoustic environments are dynamic and unpredictable. Compared to traditional solutions, our framework's ability to fine-tune signal processing parameters on the fly provides a more sustainable and scalable solution, enabling smoother deployment in industrial, automotive, and healthcare environments.

In practical terms, the application of real-world data and experiences is a key focus for future advancements. Real-world datasets, including those from challenging scenarios such as industrial noise, urban environments, and high-stress situations, will be used to further refine and enhance the system's performance. This will allow the system to continually evolve, making it more adaptable and capable of operating in increasingly complex real-world settings.

Looking forward, we foresee the continued expansion of this framework into broader domains, including next-generation AI hardware, such as intelligent hearing aids, smart headphones, and robotic audition systems. These devices will benefit from the system's real-time adaptability, which will enable them to provide more personalized, context-aware, and efficient interactions. In particular, the integration of real-world data will allow for increasingly personalized human-robot interaction, fostering a more seamless integration of AI into daily life.

Future work will focus on improving the generalization of our system across diverse real-world scenarios, including the fusion of multimodal data from acoustic, visual, and sensory sources. This will enhance the system's ability to understand and interact with the environment in ways that are not only accurate but also more human-like. Through continuous feedback loops and adaptive learning, these systems will gain a deeper understanding of their surroundings and refine their behavior accordingly, making them more effective and intuitive.

In conclusion, this paper presents a framework for integrating nonlinear acoustic models and reinforcement learning for human-robot interaction, with a focus on real-world scenarios. However, the current work is based on evaluations using existing benchmark datasets, and the true potential of this approach has yet to be fully explored in real-world environments. The next step involves simulating and utilizing real-world data to assess and refine the proposed models and methods, ensuring their adaptability and effectiveness in dynamic, uncontrolled settings. By moving beyond controlled datasets and testing in real-world conditions, we aim to push the boundaries of human-robot interaction, paving the way for more robust and responsive AI systems.

\subsection{Real-World Impact and Future Outlook}
Looking ahead, the integration of real-world data and user experiences will play a central role in shaping the future of nonlinear acoustic computing and reinforcement learning. These technologies are increasingly being designed to not only process raw data but also to interpret and understand the environment based on real-world scenarios. The future direction of AI development is fundamentally focused on how robots and AI systems can learn to observe and understand the world around them through a variety of sensory inputs, driven by the wealth of real-world data that we continuously generate.

Real-world data, whether it’s from acoustic signals, visual cues, or other sensor-based information, is essential for creating systems that interact intelligently with their environments. As AI hardware and algorithms evolve, these systems will increasingly rely on large datasets to learn from the intricacies of human behavior, environmental factors, and contextual changes. The collection and analysis of this real-world data will empower AI models to generate more accurate representations of their surroundings and adapt their responses accordingly.

In practical applications, the value of real-world data can be seen across multiple domains, from smart homes to healthcare to public safety. For instance, in the context of smart cities, AI systems will leverage large-scale sensor networks, including acoustic sensors, to continuously monitor urban environments. These systems will analyze noise levels, traffic patterns, and public safety signals to dynamically adapt city services and improve citizen quality of life. By processing data in real-time, AI systems will be able to identify patterns, optimize systems, and make decisions that reflect the immediate needs of their environment. 

Similarly, in healthcare, AI-powered diagnostic systems will draw on vast amounts of real-world patient data, learning from clinical histories, real-time physiological data, and even personal health records. These systems will improve over time by adapting to new information, offering more accurate predictions and personalized treatment recommendations. Real-world data will enable AI systems to respond to diverse medical conditions and individual patient needs, providing tailored, context-aware solutions. This experience-driven evolution of AI will be foundational to bridging the gap between theoretical models and practical, everyday use.

One of the most significant implications of real-world experience is in the field of robotics. The ability for robots to process and learn from the real-world data they collect will allow them to gain a deeper understanding of their environments. For example, robots in manufacturing plants, warehouses, and service environments will rely on data from various sensors (acoustic, visual, and haptic) to detect anomalies, interact with humans, and perform tasks in dynamic, complex settings. As robots observe their surroundings and adapt to new information, they will develop the capability to perform tasks with greater precision and adaptability, overcoming challenges such as noise interference, dynamic obstacles, and human-robot collaboration.

In these real-world settings, reinforcement learning will be important in enabling robots to improve their performance continuously. Through trial and error, robots will refine their decision-making processes based on feedback from their environment, learning to navigate increasingly complex tasks and unpredictable situations. This type of learning, grounded in real-world experience, will make robots more intuitive, responsive, and capable of independent functioning in diverse scenarios.

Looking forward, AI systems will no longer rely solely on pre-programmed rules or theoretical models; they will draw from the vast amounts of real-world experience available through data collection and user interactions. As these systems evolve, they will become increasingly capable of understanding the world in ways that mirror human cognition. For example, AI hardware like smart headphones, autonomous vehicles, and personal assistants will observe human actions, learn preferences, and adapt their responses to better align with user expectations. This data-driven adaptability will push the boundaries of human-robot interaction, creating systems that can not only react to but predict user needs, seamlessly integrating into everyday life.

Ultimately, the ability for AI systems to observe and understand the world through real-world data and experience will mark a significant shift in how machines interact with humans and their environments. The combination of nonlinear acoustic computing, reinforcement learning, and real-world data will transform the way robots and AI systems perceive the world, making them more capable, adaptable, and context-aware. As AI continues to learn from the data generated by human actions, societal trends, and environmental changes, we can expect the development of increasingly intelligent, empathetic, and responsive systems that are fully integrated into our daily lives, creating new possibilities for industries ranging from healthcare to transportation to urban management.

In conclusion, the future of AI lies in harnessing real-world data and experiences to build systems that understand and engage with the world in more human-like ways. As these technologies continue to evolve, they will redefine the boundaries of human-robot interaction, driving a new era of AI-powered solutions that enhance our daily lives, improve societal infrastructure, and open up entirely new applications across various sectors.

\appendix
\section*{Experimental Setup, Code, and Data Availability}

All code, pretrained models, noise scripts (Babble, Car, Street) and evaluation pipelines for LibriSpeech, AISHELL-1/2, Fleurs and CommonVoice are available at:
\url{https://github.com/soundai2016/nonlinear-acoustic-rl-hri}.  
Open-source databases (e.g., LibriSpeech, AISHELL-1/2, Fleurs, CommonVoice) can be downloaded from their official sites; links to each dataset and timely updates will be maintained in the repository. Detailed test-data specifications are provided in References.

\bibliography{Real_World_Nonlinear_Acoustic_Computing_Reinforcement_Learning_Human_Robot_Interaction.bib}

\end{document}